\documentclass[final]{article} 

\usepackage{amsmath,amsthm,amsfonts,amssymb,amscd}

\usepackage{thmtools}
\usepackage{mathtools}

\PassOptionsToPackage{sort&compress}{natbib}

\usepackage{hyperref}
\usepackage{url}
\usepackage{xcolor}
\usepackage{wrapfig}
\usepackage{float}
\usepackage{corl_2025,times}

\usepackage{algorithm}
\usepackage{algorithmic}

\usepackage{setspace}
\usepackage{enumitem}

\usepackage{multirow}
\usepackage{booktabs}
\usepackage{bbm}
\usepackage{svg}

\usepackage{caption}
\usepackage{subcaption}

\usepackage{array}
\usepackage{nicematrix}

\usepackage[capitalize]{cleveref}
\crefname{section}{Sec.}{Secs.}
\Crefname{section}{Section}{Sections}
\Crefname{table}{Table}{Tables}
\crefname{table}{Table}{Tables.}
\crefformat{section}{\S#2#1#3} 
\crefformat{subsection}{\S#2#1#3}
\crefformat{subsubsection}{\S#2#1#3}

\usepackage[font=small]{caption}
\usepackage{titlesec}
\titlespacing\subsubsection{0pt}{3pt plus 4pt minus 2pt}{0pt plus 2pt minus 2pt}

\makeatletter
\renewcommand\paragraph{\@startsection{paragraph}{4}{\z@}%
                                   {.2ex \@plus1ex \@minus.2ex}%
                                   {-1em}%
                                   {\normalfont\normalsize\bfseries}}
\makeatother

\colorlet{Blue}{blue!50!black}

\date{\today}  
\title{Learning Long-Context Diffusion Policies via\\Past-Token Prediction}

\author{
Marcel Torne\thanks{Equal contribution} \quad
Andy Tang\footnotemark[1] \quad
Yuejiang Liu\footnotemark[1] \quad
Chelsea Finn \\
\\
Stanford University
}

\setlist[enumerate]{leftmargin=18pt}
\setlist[itemize]{leftmargin=18pt}

\usepackage{titlesec}

\begin{document}

\maketitle


\begin{abstract}
Reasoning over long sequences of observations and actions is essential for many robotic tasks. 
Yet, learning effective long-context policies from demonstrations remains challenging. 
As context length increases, training becomes increasingly expensive due to rising memory demands, and policy performance often degrades as a result of spurious correlations.
Recent methods typically sidestep these issues by truncating context length, discarding historical information that may be critical for subsequent decisions.
In this paper, we propose an alternative approach that explicitly regularizes the retention of past information.
We first revisit the copycat problem in imitation learning and identify an opposite challenge in recent diffusion policies: rather than over-relying on prior actions, they often fail to capture essential dependencies between past and future actions.
To address this, we introduce Past-Token Prediction (PTP), an auxiliary task in which the policy learns to predict past action tokens alongside future ones.
This regularization significantly improves temporal modeling in the policy head, with minimal reliance on visual representations.
Building on this observation, we further introduce a multistage training strategy: pre-train the visual encoder with short contexts, and fine-tune the policy head using cached long-context embeddings. 
This strategy preserves the benefits of PTP while greatly reducing memory and computational overhead.
Finally, we extend PTP into a self-verification mechanism at test time, enabling the policy to score and select candidates consistent with past actions during inference.
Experiments across four real-world and six simulated  tasks demonstrate that our proposed method improves the performance of long-context diffusion policies by 3× and accelerates policy training by more than 10×.
Videos are available at \url{https://long-context-dp.github.io}.
\end{abstract}

\section{Introduction}

Many robotic tasks are inherently non-Markovian: an appropriate choice of action may depend not only on the current observation but also on past observations and actions~\citep{mandlekarWhatMattersLearning2022,zhaoLearningFineGrainedBimanual2023,leeBehaviorGenerationLatent2024,traceVLA}. For example, consider manipulation tasks where the robot arm occludes critical parts of the scene, or multi-stage tasks where early steps inform later strategies~\citep{nasirianyRoboCasaLargeScaleSimulation2024}. Likewise, past actions can prescribe a style of execution, such as speed, curvature, or strategy, that shapes how future actions should unfold~\citep{chiDiffusionPolicyVisuomotor2023,liuBidirectionalDecodingImproving2024}.

Despite the importance of historical observations, learning long-context robotic policies through imitation learning remains difficult.
First, longer observation histories often introduce features that spuriously correlate with actions in the training data. Policies that latch onto such information may diverge from expert behavior during deployment, leading to performance degradation~\citep{dehaanCausalConfusionImitation2019,ILfeedback}.
Second, conditioning on high-dimensional image sequences imposes a rapidly growing memory and computation burden, making end-to-end training excessively expensive at scale~\citep{traceVLA,liGeneralistRobotPolicies2024}.

To cope with these challenges, recent methods typically limit the amount of historical information the policy sees -- either by truncating the context length~\citep{chiDiffusionPolicyVisuomotor2023,pi0} or by engineering past observations into compact representations, such as selecting key frames~\citep{keyframeil} and summarizing observations~\citep{traceVLA}. 
While these strategies reduce memory requirement, they risk discarding information critical to subsequent decisions.

\begin{figure}[t]
    \centering
    \includegraphics[width=0.45\linewidth]{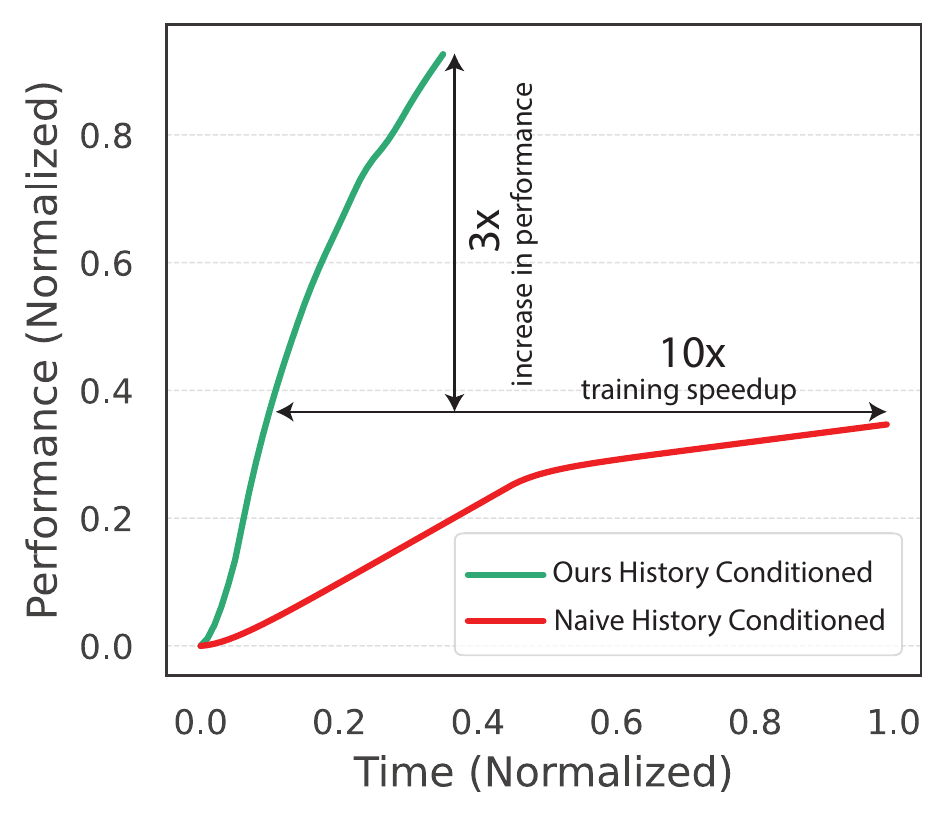}
    \includegraphics[width=0.5\linewidth]{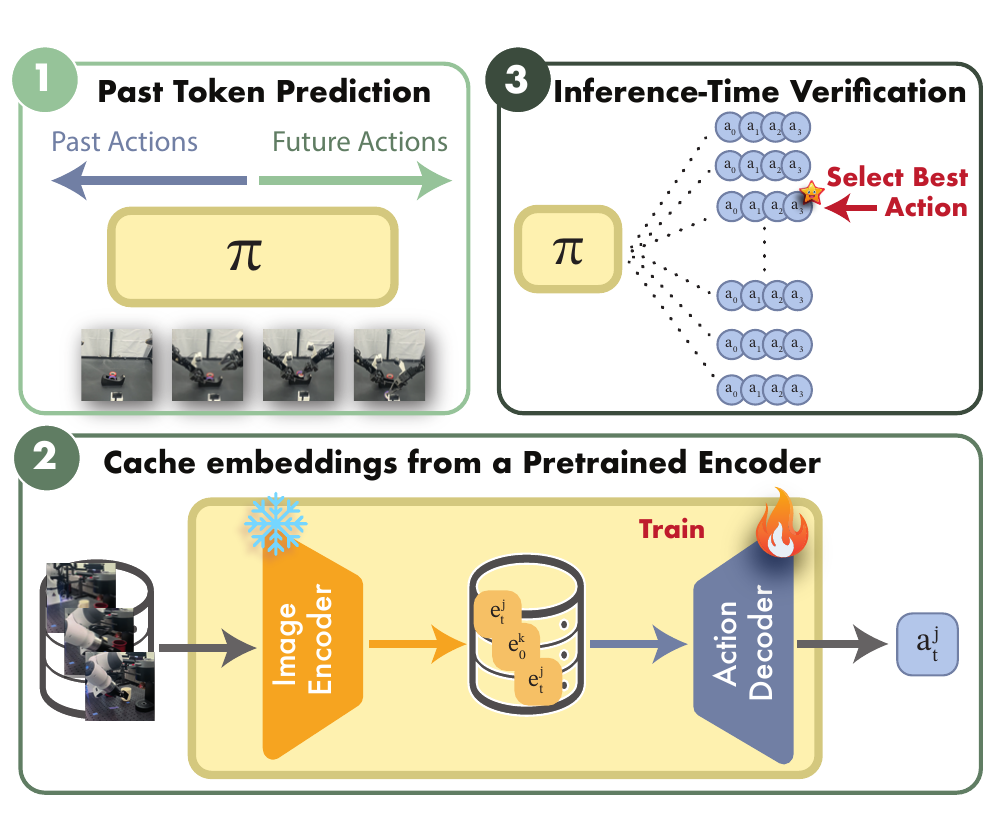}
    \caption{We propose a simple framework for learning long-context diffusion policies from human demonstrations. Our method leads to 3× gains in performance while reducing the training expense by more than 10×. }
    \label{fig:teaser}
\end{figure}

In this paper, we introduce a simple and effective approach for learning long-context robot policies, illustrated in~\cref{fig:teaser}. 
At the core of our method is to explicitly regularize the information preserved from past observations.
Specifically, we start with an analysis on the discrepancy between recent diffusion policies and their corresponding demonstrations~\citep{chiDiffusionPolicyVisuomotor2023,leeBehaviorGenerationLatent2024}.
We observe that action sequences generated by learned policies often exhibit weaker temporal dependencies than those in expert data.
To address this, we present past-token prediction (PTP), an auxiliary task where the policy learns to predict past actions alongside future ones.
This regularizer encourages the model to attend more effectively to past context, significantly boosting performance. 
Crucially, we find that the benefits of PTP primarily emerge in the policy head for sequence modeling, rather than the visual encoder.

Building upon this analysis, we introduce a multi-stage training recipe: first, pre-train the visual encoder in a short-context setting, where the policy learns to predict a chunk of future actions from only a few past frames~\citep{zhaoLearningFineGrainedBimanual2023,chiDiffusionPolicyVisuomotor2023}, and subsequently fine-tune a long-context decoder that jointly predicts past and future actions from precomputed image embeddings. This design enables the policy to capture long-range temporal dependencies while substantially reducing memory and computational overhead.
Beyond training, we further leverage PTP as a self-verification mechanism during inference. 
At each time step, the policy generates multiple candidate actions and selects the one most consistent with its previously executed actions.

In summary, our main contributions are twofold: (i) identify a critical discrepancy in temporal action dependencies between learned policies and expert demonstrations (\cref{sec:preliminary}), (ii) propose a training and inference method for long-context imitation learning via past-token prediction (\cref{sec:method}).
Empirically, we validate our method on diffusion-based policies~\citep{chiDiffusionPolicyVisuomotor2023} across six simulation and four real-world tasks (\cref{sec:experiment}).
On average, our method increases the success rate of long-context policies by 3× while reducing training overhead by over 10 times.
Notably, it enables policies to achieve 80\% success on history-critical tasks where existing methods fail entirely.

\section{Related Work}

\paragraph{Imitation Learning.}
Imitation learning has long served as a simple yet powerful paradigm for robot learning~\citep{argallSurveyRobotLearning2009,ravichandarRecentAdvancesRobot2020,zareSurveyImitationLearning2024}. Early approaches typically framed it as a supervised learning problem, where the policy learns to map a given observation to the target action~\citep{rossEfficientReductionsImitation2010a}. 
More recent works have shifted toward modeling the distribution of demonstrations~\citep{zhaoLearningFineGrainedBimanual2023,chiDiffusionPolicyVisuomotor2023,leeBehaviorGenerationLatent2024,ze3DDiffusionPolicy2024,bharadhwajRoboAgentGeneralizationEfficiency2024,wangEquivariantDiffusionPolicy2024,haldarBAKUEfficientTransformer2024,liuBidirectionalDecodingImproving2024}. 
This approach has recently achieved remarkable success towards generalist robot policies~\citep{brohanRT1RoboticsTransformer2023,black$p_0$VisionLanguageActionFlow2024}. However, imitation learning remains highly susceptible to covariate shift \citep{dehaanCausalConfusionImitation2019, wenFightingCopycatAgents2020,shaoUnifyingFrameworkCausal2025}, e.g. \citet{dagger} and \citet{ILfeedback} characterize compounding errors in a feedback loop once the learned policy diverges from the demonstration manifold. This problem is exacerbated by high-dimensional visual inputs, where less robust features might be learned due to underspecification~\citep{rt-affordance}.
Notably, recent works~\citep{chiDiffusionPolicyVisuomotor2023,traceVLA} have empirically found that image-conditioned specialist and generalist policies degrade with history, leading many works to exclude history altogether~\citep{teamOctoOpenSourceGeneralist2024, black$p_0$VisionLanguageActionFlow2024, brohanRT2VisionLanguageActionModels2023,zhaoLearningFineGrainedBimanual2023,kimOpenVLAOpenSourceVisionLanguageAction2024,torneRobotLearningSuperLinear2024,liHAMSTERHierarchicalAction2024}. Our work introduces and analyzes a training recipe that counteracts this degradation.

\paragraph{Long-Context Policies.}
Handling long sequences of high-dimensional observations has been a persistent challenge in robot learning. A common strategy is to reduce the input history—by discarding parts of the past via adversarial regularization~\citep{wenFightingCopycatAgents2020}, information bottlenecks~\citep{PALR}, or selecting salient subsets through techniques like keyframes~\citep{keyframeil} and motion tracks~\citep{motiontracksil}. Other methods construct higher-level summaries, such as sketch synthesis~\citep{rt-sketch} or visual trace prompting~\citep{traceVLA}, especially for generalist policies.
These approaches rely on the assumption that much of the historical context is irrelevant—a simplification that may break down in temporally complex tasks.
An alternative line of work attempts to model the full context in an autoregressive manner using action tokens~\citep{radosavovicRobotLearningSensorimotor2023,fuInContextImitationLearning2024}.
Yet, designing action tokenizers that can effectively capture long-range temporal structure remains an open problem~\citep{vuongActionTokenizerMatters2025}.
Our method takes an orthogonal approach: we explicitly regularize diffusion policies to retain information about past actions that would otherwise be lost from historical context.


\paragraph{Test-Time Scaling.}
Recent research in language modeling, image generation, and robotics has shown that inference-time compute may allow models to improve their performance \citep{bansalSmallerWeakerBetter2024a,maInferenceTimeScalingDiffusion2025,nakamotoSteeringYourGeneralists2024}. Some seek to build an additional verifier to re-rank the output samples~\citep{cobbeTrainingVerifiersSolve2021,wengLargeLanguageModels2023,lightmanLetsVerifyStep2023,yuOVMOutcomesupervisedValue2024}, while others propose to leverage the internal knowledge to improve reasoning through self-verification \citep{stechlySelfVerificationLimitationsLarge2024}.
Our method echoes the latter paradigm in the robotic context: our policy is trained to predict accurate past actions before predicting the present action and can self-verify at test-time through past action accuracy. Similarly to how it may be more compute-efficient to use test-time compute on a small LLM \citep{llmscaling}, we show that checkpoints trained for fewer epochs or at shorter histories can approach the performance of optimal checkpoints by using more test-time compute.

\section{Preliminaries}
\label{sec:preliminary}

\paragraph{Problem Setting.}

We consider the problem of imitation learning, where a robot learns to perform complex tasks from expert demonstrations. At each time step $t$, the robot receives a visual observation $o_t$ and executes an action $a_t$. Crucially, we assume that each observation $o_t$ contains only partial information about the underlying state $s_t$, but the complete information about $s_t$ can be inferred from the history of observations. This setting encapsulates practical challenges commonly encountered in robotic tasks, such as latent strategies in the demonstrations (e.g., expert preference), temporal context (e.g., stage within a task), and perceptual limitations (e.g., visual occlusions).

Given a dataset of $N$ expert demonstrations $\mathcal{D} = \{\tau_i\}_{i=1}^{N}$, where each demonstration trajectory $\tau_i$ consists of a sequence of observation-action pairs, our goal is to learn a long-context policy 
$\pi_\theta(\mathbf{a}_{t:t+l} | \mathbf{o}_{t-k:t})$ that takes as input the current observation along with the history $\mathbf{o}_{t-k:t} = (o_{t-k}, \dots, o_t)$ over the past $k$ time steps, and predicts the current and future actions $\mathbf{a}_{t:t+l} = (a_t, \dots, a_{t+l})$ spanning the next $l$ time steps.
While increasing the context length $k$ provides richer historical information, the resulting long-context policies often suffer from substantial performance  declines~\citep{chiDiffusionPolicyVisuomotor2023,traceVLA}.

\paragraph{Practical Challenges.}
One central challenge in long-context imitation learning arises from the prevalence of spurious features in observation history. 
As context length increases, the model is exposed to a growing set of input features, some of which correlate with but do not causally influence the expert actions. Policies relying on these spurious features in observation history may reach high prediction accuracy within the training distribution but generalize poorly during deployment~\citep{dehaanCausalConfusionImitation2019}. 
One notable manifestation is the copycat behavior~\citep{wenFightingCopycatAgents2020}, where the learned policy simply mimics previous actions as predictions for future ones, ignoring current state observations.
Does this phenomenon persist in modern imitation learning methods?

To understand this, we evaluate temporal action dependencies by measuring how predictable the current action is from prior actions alone.
Specifically, given a set of demonstrations, we first train long-context policies with varying observation history lengths.
We then collect policy rollouts and train a simple two-layer MLP $\phi(a_t | a_{t-1})$ to predict the current action based solely on the previous action.
We measure the mean-squared error $\epsilon_{\pi}$ of the MLP predictor on holdout rollouts and similarly obtain $\epsilon_{\pi^*}$ for expert demonstrations.
Following~\citep{wenFightingCopycatAgents2020,seoRegularizedBehaviorCloning2023}, we define the action predictability ratio as $\epsilon_{\pi^*} / \epsilon_{\pi}$. 
Intuitively, a ratio greater than 1 indicates an over-reliance on previous actions (i.e., copycat behavior), while a ratio less than 1 indicates weaker-than-expert action dependency. 

\begin{figure}[t]
    \centering
    \begin{minipage}[t]{0.35\linewidth}
        \centering
        \includegraphics[width=\linewidth]{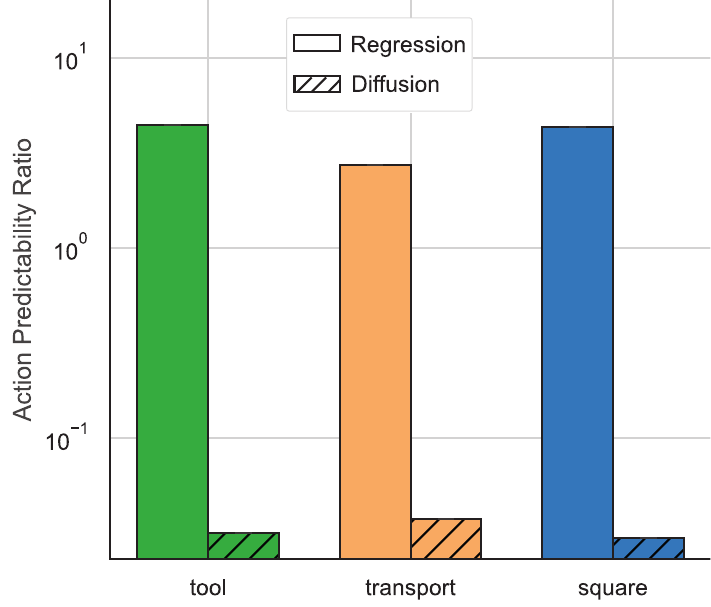}
        \caption{Comparison of regression-based and diffusion-based policies in temporal action dependency, normalized by that in demonstrations. 
        }
        \label{fig:ptp_correlationanalysis}
    \end{minipage}\hfill
    \begin{minipage}[t]{0.615\linewidth}
        \centering
        \includegraphics[width=\linewidth]{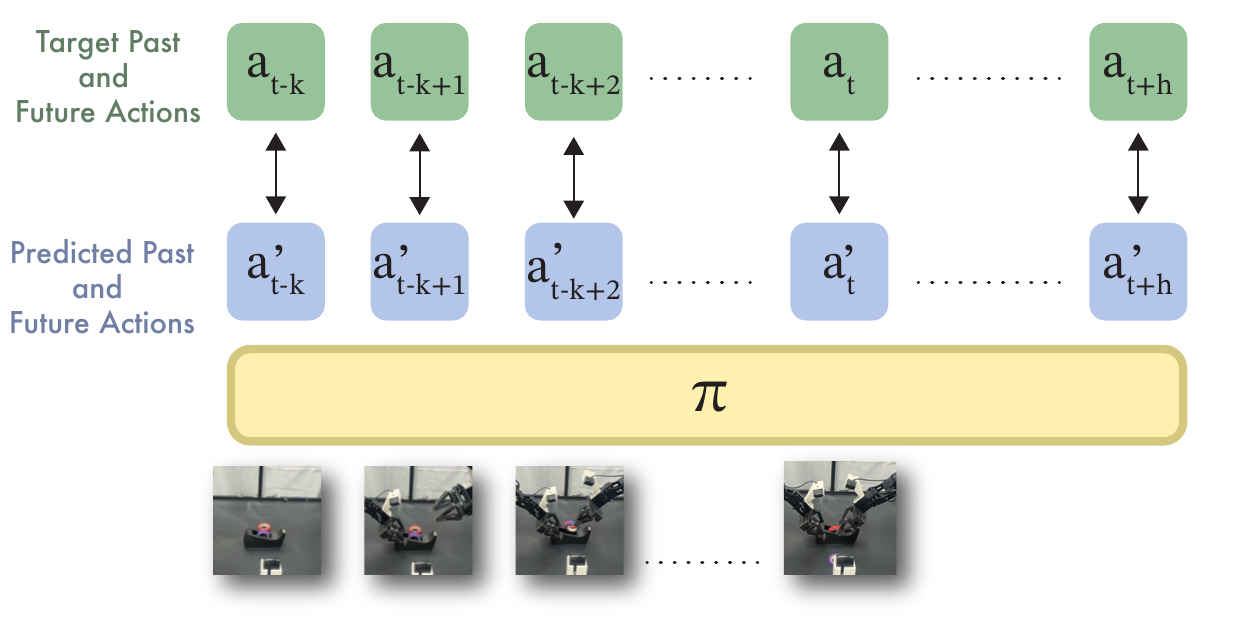}
        \caption{Illustration of past-token prediction. The policy head is trained to jointly predict both past and future action tokens, encouraging the model to capture the temporal dependencies that are otherwise lost between past and future actions.}
        \label{fig:ptp_illustration}
    \end{minipage}
\end{figure}

\cref{fig:ptp_correlationanalysis} shows the action predictability ratios for classical regression-based policies and modern diffusion-based policies~\citep{chiDiffusionPolicyVisuomotor2023} across three simulation tasks in RoboMimic~\citep{mandlekarWhatMattersLearning2022}.
Interestingly, the two approaches exhibit opposite failure modes:
The regression-based policies indeed exhibit high action predictability, even exceeding that of the expert demonstrations.
In contrast, {\it modern diffusion-based policies yield predictability ratios significantly below 1, indicating a surprising underuse of past action information despite conditioning on long observation histories.}
Ideally, an effective imitator should not only learn to accurately predict expert actions in the training set, but also reach a similar level of temporal action dependencies in its rollouts. We will next introduce a method designed to explicitly bridge this gap.

\section{Method}
\label{sec:method}

In this section, we introduce a long-context imitation learning method, aiming to improve both policy performance and training efficiency.
We will first describe a simple but crucial auxiliary task to enhance temporal dependencies in sequential decision-making (\cref{sec:pap}). 
We will then present a multi-stage training recipe that preserves the benefit of this auxiliary task while reducing memory consumption (\cref{sec:cache}). Finally, we will introduce an inference technique that leverages the auxiliary task to effectively self-verify sampled predictions at test time (\cref{sec:tt_method}).

\subsection{Past-Token Prediction}
\label{sec:pap}

One common design choice in imitation learning is next-token prediction, where the policy predicts only the immediate next action token at each time step. To better capture temporal dependencies, recent methods have extended this to predict a chunk of future action tokens~\citep{zhaoLearningFineGrainedBimanual2023,chiDiffusionPolicyVisuomotor2023}. However, as shown in \cref{sec:preliminary}, this design alone remains insufficient for modeling the critical dependencies between past and future decisions.

We address this issue through Past-Token Prediction (PTP), an auxiliary objective that tasks the policy to predict past action tokens alongside future ones. Formally, given a sequence of observations $\mathbf{o}_{t-k:t}$, the policy is trained to jointly predict the action tokens from the past time step $t-k$ to the upcoming time step $t + h$:
\begin{equation}
\hat{\mathbf{a}}_{t-k:t+h} = \pi_{\theta}(\mathbf{o}_{t-k:t}).
\end{equation}
As illustrated in~\cref{fig:ptp_illustration}, this objective expands the prediction window in both temporal directions, explicitly encouraging the policy to preserve information about past actions from the history context. 

\subsection{Memory-Efficient Training with PTP}
\label{sec:cache}

Recent imitation learning approaches typically train visuomotor policies end-to-end, jointly optimizing both the visual encoder and the policy head. However, this strategy incurs memory costs that grow linearly with context length, making it prohibitively expensive to train long-context policies.

To address this, we propose a multi-stage training recipe that decouples visual representation learning from policy optimization. Our training process consists of three specific stages:
\begin{wrapfigure}{r}{0.525\linewidth}
    \centering
    \includegraphics[width=\linewidth]{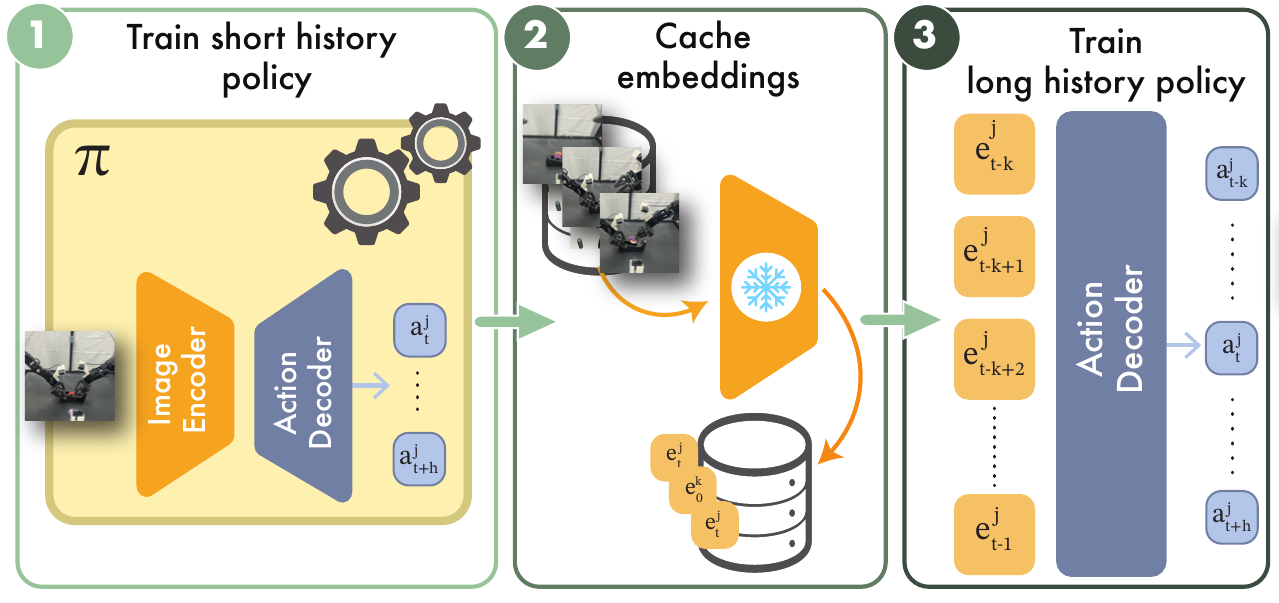}
    \caption{Overview of multistage training with embedding caching. As PTP acts on the decoder, caching embeddings substantially improves inference speed without sacrificing performance. We use a visual encoder from a short-range policy with low validation loss to compute the embeddings of the images in the buffer and cache them in the buffer. With the cached embeddings we can train the long-horizon policy much faster. At test time we take the original encoder.}
    \label{fig:cache_emb_method}
    \vspace{-10pt}
\end{wrapfigure}
\begin{enumerate}
    \vspace{-10pt}
    \item \textbf{Encoder Training:} We first train the visual encoder with a short observation context but a long prediction horizon, encouraging it to extract representations that retain information critical for predicting $l$ subsequent steps.
    \item \textbf{Feature Caching:} We then freeze the encoder and precompute embeddings for all frames in the training set. This caching step eliminates redundant computation during policy training.
    \item \textbf{Policy Training:} Finally, we train the policy head conditioned on long-context observations represented by the cached embeddings. This enables the policy to model long-range dependencies without repeatedly processing visual inputs. 
\end{enumerate}

As shown in~\cref{fig:cache_emb_method}, this multistage training approach retains a computational footprint similar to short-context training while enabling efficient scaling to longer observation contexts. In Appendix \ref{apdx:effectinternal}, we show in more detail how the features of a short-history policy are sufficient to support strong long-context performance.

\subsection{Test-Time Verification with PTP}
\label{sec:tt_method}

\begin{wrapfigure}{r}{0.525\linewidth}
    \centering
    \vspace{-15pt}  
    \includegraphics[width=\linewidth]{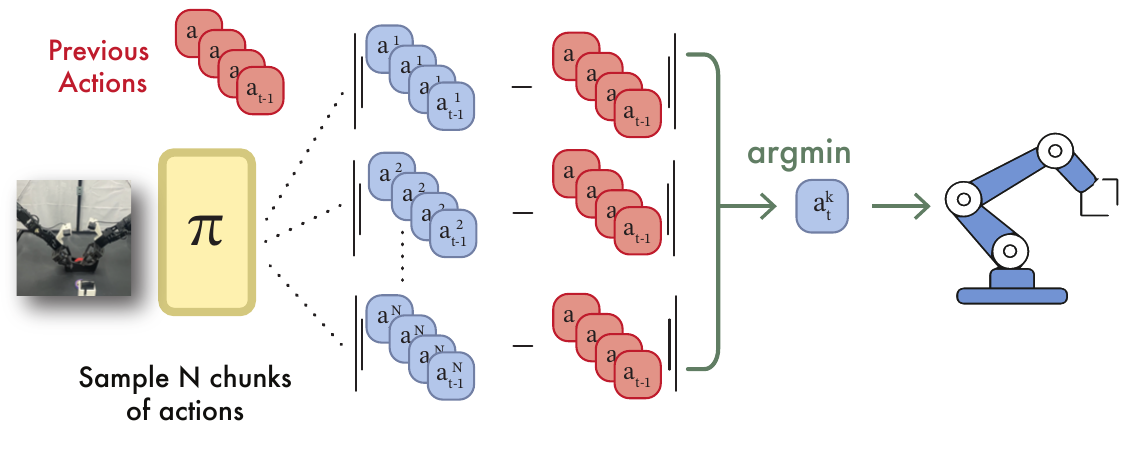}
    \caption{Test-time verification. Multiple action sequences are sampled from the same observation, and the policy selects the sequence that is most consistent compared to ground-truth previous actions.}
    \label{fig:test-time}
    \vspace{-10pt}  
\end{wrapfigure}
Another common challenge in recent diffusion policies lies in the robustness of sampled predictions. Often, not all samples are equally good at capturing the critical temporal dependencies. 
Recent work has explored re-ranking sampled predictions based on consistency with past predictions~\citep{liuBidirectionalDecodingImproving2024}.
However, when the previous prediction for future actions is suboptimal, e.g. because of unexpected environmental changes, this approach may propagate errors rather than correct them.

To address this shortcoming, we cast Past-Token Prediction as a self-verification mechanism during deployment.
At each inference step, we sample a batch of $B$ candidate action sequences:
\begin{equation}
\mathcal{A} = \{\hat{\mathbf{a}}^{(1)}, \dots, \hat{\mathbf{a}}^{(B)}\}, \quad \hat{\mathbf{a}}^{(i)} \sim \pi_\theta(\mathbf{o}_{t-k:t}),
\end{equation}
where each sampled candidate $\hat{\mathbf{a}}^{(i)} = (a_{t-k}, \dots, a_{t+h})^{(i)}$ includes both reconstructed past actions and predicted future actions. Since the first $k-1$ actions have already been executed, we use them as a ground-truth reference and select the candidate whose reconstructed past actions best match the executed ones:
\begin{equation}
\hat{\mathbf{a}}^* = \arg\min_{\hat{\mathbf{a}} \in \mathcal{A}} \sum_{\tau=t-k}^{t-1} \|\hat{a}_{\tau} - a_{\tau}\|^2
\end{equation}
As illustrated in~\cref{fig:test-time}, this sample selection procedure is fully parallelizable on GPU devices, enabling self-verification of temporal action dependencies with minimal computational overhead.  

\section{Experiments}
\label{sec:experiment}

In this section, we evaluate the proposed method for learning long-context diffusion policies. We seek to answer the following questions regarding policy performance and training efficiency:
\begin{enumerate}[nosep]
    \item How effectively does PTP mitigate the lack of temporal action dependencies shown in \cref{sec:preliminary}?
    \item How well do the resulting policies perform on tasks that require history-aware decision-making?
    \item To what extent does the proposed multi-stage training recipe accelerate policy learning?
    \item Could PTP verification further mitigate deficiencies in temporal dependencies at test time?
    \item Finally, how do these findings generalize to history-critical tasks in the real world?
\end{enumerate}

\begin{figure}[t]
    \centering
    \begin{minipage}[t]{0.32\linewidth}
        \centering
        \includegraphics[width=\linewidth]{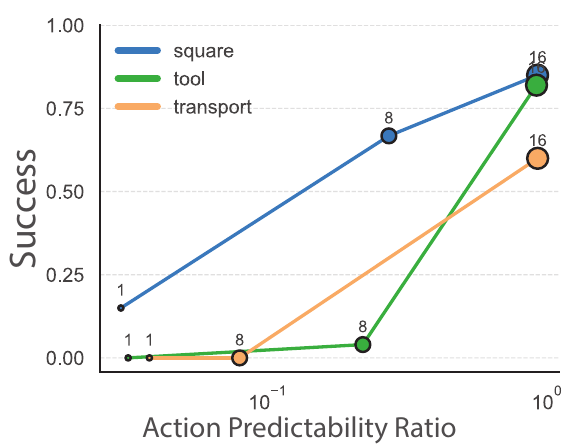}
        \caption{Effect of PTP on temporal action dependency and policy performance. Increasing the amount of past-token supervision aligns the learner more closely with expert action dependencies, resulting in higher success rates.}
        \label{fig:ptp_correlation}
    \end{minipage}\hfill
    \begin{minipage}[t]{0.32\linewidth}
        \centering
        \includegraphics[width=\linewidth]{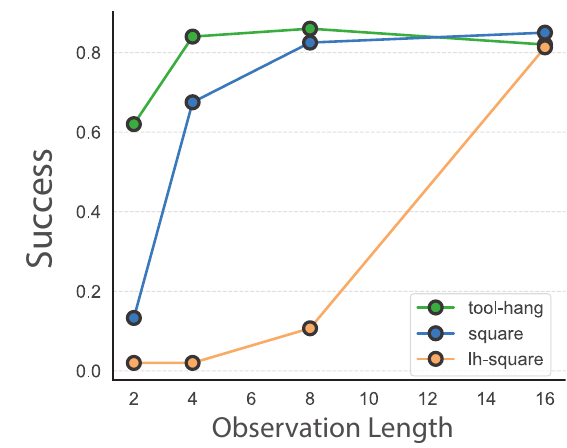}
        \caption{Effect of history observations on PTP-trained diffusion policies. Increasing the context length progressively enhances policy performance, especially in history-critical tasks such as Long-Horizon Square}
        \label{fig:ptp_history}
    \end{minipage}\hfill
    \begin{minipage}[t]{0.32\linewidth}
        \centering
        \includegraphics[width=\linewidth]{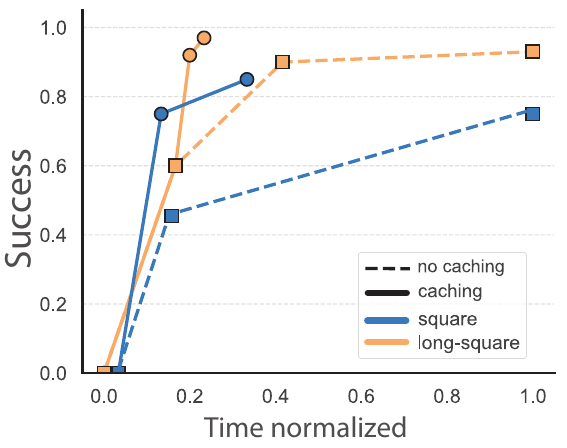}
        \caption{Effect of feature caching. Caching speeds up training by over 5× without hurting performance. On complex tasks like Tool Hang, long-context policies fail to perform even after two days without caching.}
        \label{fig:cach}
    \end{minipage}
\end{figure}

To this end, we evaluate our method on the modern diffusion-based policy~\citep{chiDiffusionPolicyVisuomotor2023}, in comparison with the classical regression-based policy.
By default, both policies receive visual and proprioceptive observations from the past 16 time steps as conditional input.
We compare policies trained with {\it PTP} against two baselines: {\it no-history} policies that take only the current and past single frame as input, and {\it no-PTP} policies that are trained without PTP.
Unless otherwise specified, all policies are trained using the multistage recipe with feature caching and evaluated under a single-sample inference setting. The effect of test-time verification is evaluated separately across multiple checkpoints under varying sample budgets.
Additional results are presented in~\cref{apdx:experiment}, with implementation details provided in~\cref{apdx:envanddata}.

\subsection{Simulation Experiments}
\label{sec:simexp}

We first evaluate our method across six simulated tasks. Four of these are sourced from existing benchmarks: \emph{square, tool hang, and transport} from RoboMimic~\citep{mandlekarWhatMattersLearning2022}, each provided by multi-human demonstration datasets, and Push-T from~\citet{chiDiffusionPolicyVisuomotor2023}.
These tasks feature diverse strategies in demonstrations, requiring the policy to infer and commit to consistent behaviors over time based on historical context.
In addition, we introduce two new long-horizon simulation tasks:
\emph{long-horizon square}, where the robot must place and remove a square onto the peg twice before finally dropping it in the peg; and \emph{long-horizon aloha}, where one arm must pick up a block, move it to the center of the field of view, and return it precisely to its original location.
Success in these new tasks critically depends on the ability to recall and act upon information observed earlier in the episode.
Each policy-task pair is evaluated over 100 episodes across three random seeds.
We next summarize the key findings from these simulation experiments.

\textbf{\emph{Takeaway 1: PTP mitigates deficiencies in modeling temporal action dependencies.}}
To validate the effect of PTP on modeling temporal action dependencies, we use the same set of tasks as in~\cref{sec:preliminary} and train policies to predict a variable number of past tokens $\{\hat{a}_{t-c-1}, \dots, \hat{a}_t\}$, where $c$ denotes the number of actions included in the prediction target.
Specifically, we compare three variants: (i) {\it no-PTP} with $c=1$, equivalent to the vanilla next-token prediction baseline; (ii) {\it half-PTP} with $c=8$, which predicts action tokens corresponding to half the observation window; and (iii) {\it full-PTP} with $c=16$. 
As shown in~\cref{fig:ptp_correlation}, PTP consistently increases the action predictability and gets closer to that observed in the expert demonstrations. 
Notably, the non-PTP baseline exhibits approximately 10× to 100× weaker action predictability ratios compared to expert behavior, whereas full-PTP yields temporal dependencies comparable to demonstrations.

\begin{figure}
    \centering
    
    \includegraphics[width=0.99\textwidth]{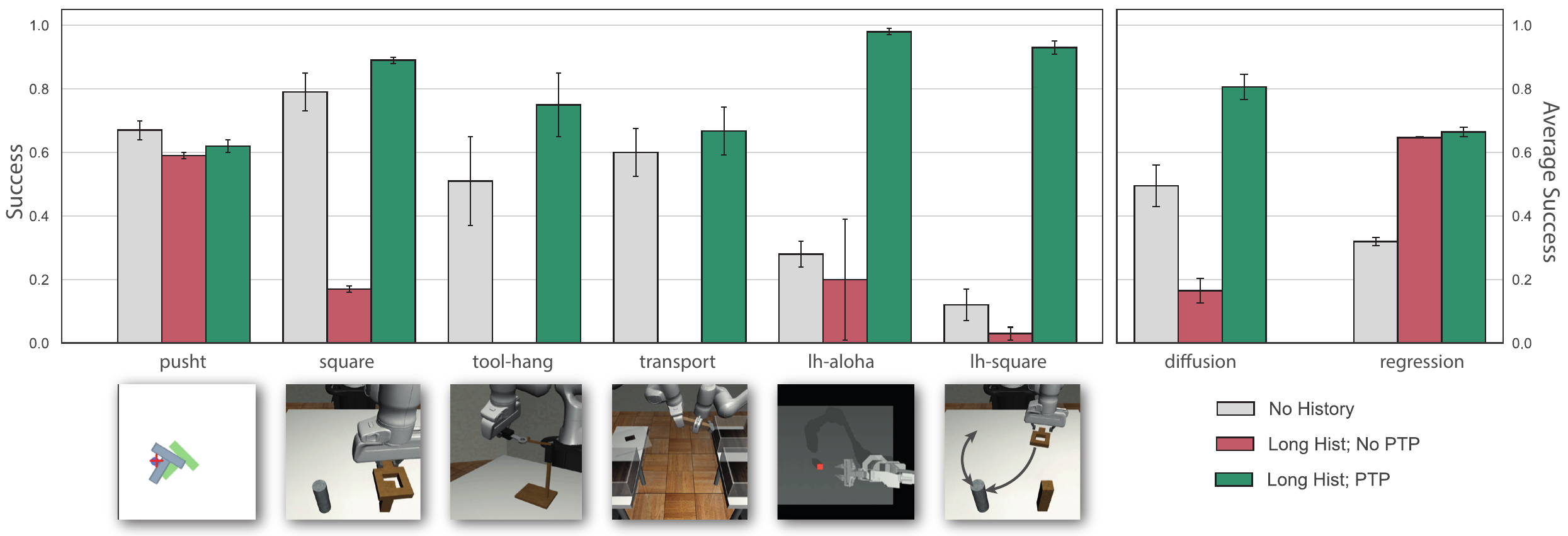}

    \caption{Comparison of different policies across six simulation tasks. Unlike classical regression-based policies, modern diffusion-based policies exhibit a clear drop in performance when conditioned on historical observations.
    Our method achieves an average improvement of over 30\% compared to no-history diffusion policies, and over 60\% compared to no-PTP diffusion policies. The gains are especially pronounced on history-critical tasks such as \emph{long-horizon aloha} and \emph{long-horizon square}.
    }
    \label{fig:mainsimresults}
\end{figure}

\textbf{\emph{Takeaway 2: PTP significantly improves the performance of modern policies.}} 
To assess the impact of PTP on task performance, we compare our method against the no-history and no-PTP baselines on two classes of policies: diffusion-based versus regression-based.
All models are evaluated following the protocol from~\citep{chiDiffusionPolicyVisuomotor2023}, with action chunking set to 8 time steps.
As shown in~\cref{fig:mainsimresults}, while the {\it no-history} baseline already performs competitively on some existing tasks, PTP matches or surpasses its performance. The advantage of PTP is particularly pronounced in long-horizon tasks: both the {\it no-history} and {\it no-PTP} baselines struggle with success rates below 30\%, whereas our method achieves near-perfect performance on the long-horizon tasks. 
Averaged across all six simulation tasks, PTP yields an average 50\% improvement for diffusion-based policies when conditioned on long contexts, and outperforms the regression counterpart by nearly 20\%.

\textbf{\emph{Takeaway 3: PTP-trained policies benefit from longer contexts.}}
To further understand the role of historical contexts, we evaluate PTP-trained diffusion-based policies conditioned on observation histories of varying lengths, ranging from 2 to 16 time steps. As shown in ~\cref{fig:ptp_history}, longer histories generally lead to improved performance.
For relatively simple tasks such as \emph{square}, gains tend to saturate beyond 4 steps; however, for more complex tasks, such as \emph{transport}, \emph{long-horizon square}, and \emph{long-horizon aloha}, longer contexts provide substantial performance boosts.

\textbf{\emph{Takeaway 4: Embedding caching accelerates PTP training without sacrificing performance.}}
To assess the effectiveness of the proposed multistage training strategy, we train history-conditioned diffusion policies with and without embedding caching for two days on the three tasks used above (\cref{sec:preliminary}), evaluating checkpoints saved every 50 epochs.
As shown in~\cref{fig:cach}, the vanilla training recipe without caching completes only a limited number of epochs within the time budget. In contrast, our caching-based approach matches performance in just 20\% of the training time and surpasses it within 40\% of the compute budget.

\begin{figure}[t]
    \centering
    \begin{minipage}[t]{0.4\linewidth}
        \centering
        \includegraphics[width=0.7\linewidth]{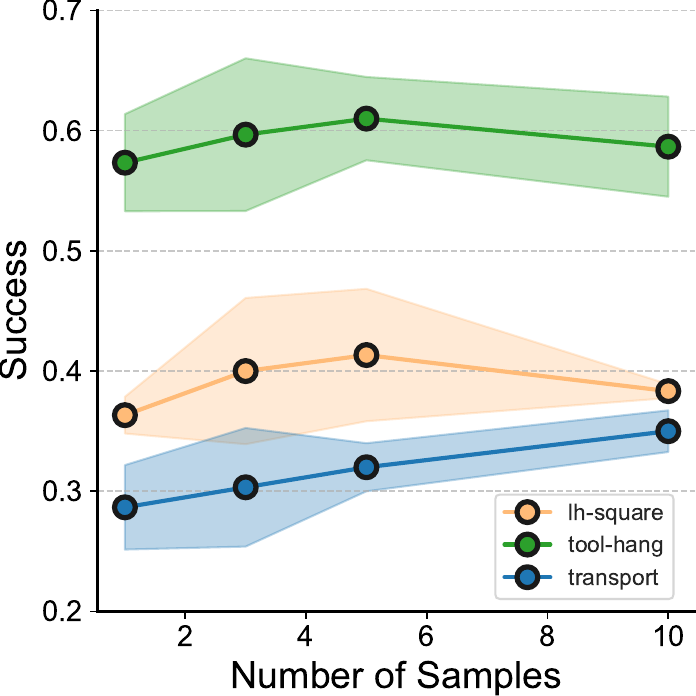}
        \caption{Effect of PTP self-verification. Increasing sampling budgets yields a 5\% gain in challenging closed-loop settings.}
        \label{fig:testtime_update}
    \end{minipage}
    \hfill
    \begin{minipage}[t]{0.58\linewidth}
        \centering
        \includegraphics[width=\linewidth]{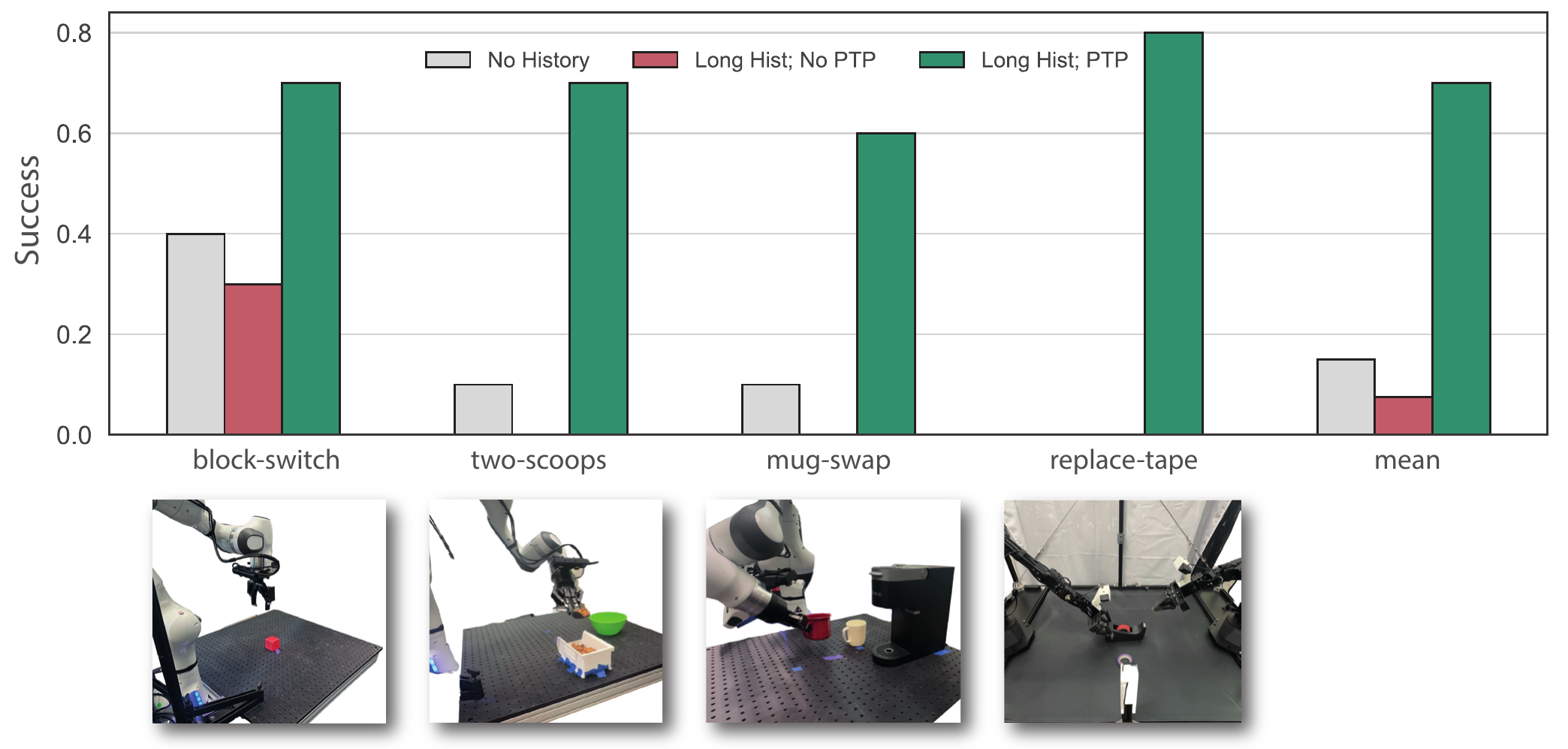}
        \caption{Comparison of different policies on four real-world tasks that critically depend on historical context. Our method yields over a 55\% improvement compared to baselines.}
        \label{fig:realworld_results}
    \end{minipage}
\end{figure}

\textbf{\emph{Takeaway 5: PTP verification boosts performance in challenging settings at test time.}} 
To validate the potential of self-verification through PTP, we evaluate history-conditioned policies on three challenging tasks, including Tool Hang, Transport, and Long Square, trained under constrained compute budgets and tested with varying sampling budgets \{1, 3, 5, 10\}.
As shown in \cref{fig:testtime_update}, PTP-guided sample selection provides notable performance gains. Notably, increasing the number of sampled candidates from 1 to 5 results in approximately 5\% improvement in success rate on all these tasks.

\subsection{Real-World Experiments}

We next examine our method on four history-critical tasks across two robot platforms in the real world: \emph{Franka block switch}: move a block from one side to another, where history is needed to correctly infer which side to place the block; \emph{Franka two scoops}, transport two scoops to the target, where history is needed to count scoops; \emph{Franka mug replacement} and \emph{ALOHA tape replacement}: replace one mug or tape by another, where history is needed to distinguish old and new objects.
Across all tasks, we use diffusion-based policies with a context length of 16 and a chunk size of 8. Due to different ranges of temporal dependency in these tasks, we apply task-specific subsampling rates detailed in~\cref{apdx:envanddata}.

\textbf{\emph{Quantitatively, PTP outperforms baselines by over 4× in the real world.}}
As shown in~\cref{fig:realworld_results}, the {\it no-history} baseline is limited to an average success rate of 15\% due to the absence of critical history information. 
The {\it no-PTP} baseline, which simply conditions on history without PTP, yields near-zero success on three of four tasks. In contrast, our method achieves an average 70\% success rate. Notably, on Tape Replacement, one of the most challenging tasks across the board, our method achieves 80\% success, while the two baselines fail entirely.

\textbf{\emph{Qualitatively, PTP-trained long-context policies excel at both high-level and low-level memory.}}
As shown in the videos on the website, the two baselines exhibit distinct failure modes: 
the {\it no-history} policies often fail at high-level decision-making, such as replacing the wrong object or miscounting scoops, whereas the {\it no-PTP} baseline struggles with low-level motor control, such as unsuccessful grasps and inaccurate placements. In comparison, policies trained with our method demonstrate improvement in both high-level planning and low-level control, resulting in more coherent and reliable behavior across tasks.

\section{Conclusion}

We have presented Past Token Prediction (PTP), a simple yet effective auxiliary objective for learning history-conditioned diffusion policies from demonstrations. 
We have shown that PTP can effectively strengthen temporal action dependencies that are often lost in recent diffusion policies.
In addition, we have introduced a multistage training strategy and a self-verification mechanism that allow for effective use of PTP during both training and inference.
Extensive experiments across ten manipulation tasks in both simulations and the real world demonstrate its advantages in efficiency and effectiveness. 


\section{Limitations and Discussion.} 

Our work has focused on extending context length specifically for diffusion policies, motivated by their growing prevalence in the robot learning community.
Nevertheless, the effectiveness of our method may generalize to other classes of modern policies as well. In fact, concurrently with our work,~\citet{vuongActionTokenizerMatters2025} observes similar challenges in tokenization-based policies. Extending our approach to such settings, and more broadly, designing action tokenizers that better preserve temporal structure, can be an exciting avenue for future research.

Another practical challenge our method faces is inference overhead. While we have shown that caching and reusing visual embeddings can substantially reduce memory consumption and speed up policy training, inference overhead remains a practical bottleneck for closed-loop operations. 
To make inference time manageable, we followed common practices from recent literature by downsampling observation history and extending action chunk. However, these adjustments are known to compromise policy reactivity. Designing strategies to further accelerate inference—particularly given the growing scale of VLA models—could be another fruitful direction for future research.

\section*{Acknowledgments}
We thank Sergey Levine and the members of the Stanford IRIS lab for their insightful discussions. We thank Joey Hejna, Moo Jin Kim, Alec Lessing, and Amber Xie for their support with the real world experiments. This work was supported in part by The Robotics and AI Institute, DARPA, and the ONR grant N00014-22-1-2621. C. Finn is a CIFAR fellow. Y. Liu is an SNSF postdoc fellow.


\begin{thebibliography}{48}
\providecommand{\natexlab}[1]{#1}
\providecommand{\url}[1]{\texttt{#1}}
\expandafter\ifx\csname urlstyle\endcsname\relax
  \providecommand{\doi}[1]{doi: #1}\else
  \providecommand{\doi}{doi: \begingroup \urlstyle{rm}\Url}\fi

\bibitem[Argall et~al.(2009)Argall, Chernova, Veloso, and Browning]{argallSurveyRobotLearning2009}
Brenna~D. Argall, Sonia Chernova, Manuela Veloso, and Brett Browning.
\newblock A survey of robot learning from demonstration.
\newblock \emph{Robotics and Autonomous Systems}, 57\penalty0 (5):\penalty0 469--483, May 2009.

\bibitem[Bansal et~al.(2024)Bansal, Hosseini, Agarwal, Tran, and Kazemi]{bansalSmallerWeakerBetter2024a}
Hritik Bansal, Arian Hosseini, Rishabh Agarwal, Vinh~Q. Tran, and Mehran Kazemi.
\newblock Smaller, {{Weaker}}, {{Yet Better}}: {{Training LLM Reasoners}} via {{Compute-Optimal Sampling}}.
\newblock \emph{arXiv preprint arXiv:2408.16737}, August 2024.

\bibitem[Bharadhwaj et~al.(2024)Bharadhwaj, Vakil, Sharma, Gupta, Tulsiani, and Kumar]{bharadhwajRoboAgentGeneralizationEfficiency2024}
Homanga Bharadhwaj, Jay Vakil, Mohit Sharma, Abhinav Gupta, Shubham Tulsiani, and Vikash Kumar.
\newblock {{RoboAgent}}: {{Generalization}} and {{Efficiency}} in {{Robot Manipulation}} via {{Semantic Augmentations}} and {{Action Chunking}}.
\newblock In \emph{2024 {{IEEE International Conference}} on {{Robotics}} and {{Automation}} ({{ICRA}})}, pages 4788--4795, May 2024.

\bibitem[Black et~al.(2024{\natexlab{a}})Black, Brown, Driess, Esmail, Equi, Finn, Fusai, Groom, Hausman, Ichter, Jakubczak, Jones, Ke, Levine, {Li-Bell}, Mothukuri, Nair, Pertsch, Shi, Tanner, Vuong, Walling, Wang, and Zhilinsky]{black$p_0$VisionLanguageActionFlow2024}
Kevin Black, Noah Brown, Danny Driess, Adnan Esmail, Michael Equi, Chelsea Finn, Niccolo Fusai, Lachy Groom, Karol Hausman, Brian Ichter, Szymon Jakubczak, Tim Jones, Liyiming Ke, Sergey Levine, Adrian {Li-Bell}, Mohith Mothukuri, Suraj Nair, Karl Pertsch, Lucy~Xiaoyang Shi, James Tanner, Quan Vuong, Anna Walling, Haohuan Wang, and Ury Zhilinsky.
\newblock \${$\pi\_$}0\$: {{A Vision-Language-Action Flow Model}} for {{General Robot Control}}.
\newblock \emph{arXiv preprint arXiv:2410.24164}, November 2024{\natexlab{a}}.

\bibitem[Black et~al.(2024{\natexlab{b}})Black, Brown, Driess, Esmail, Equi, Finn, Fusai, Groom, Hausman, Ichter, Jakubczak, Jones, Ke, Levine, Li-Bell, Mothukuri, Nair, Pertsch, Shi, Tanner, Vuong, Walling, Wang, and Zhilinsky]{pi0}
Kevin Black, Noah Brown, Danny Driess, Adnan Esmail, Michael Equi, Chelsea Finn, Niccolo Fusai, Lachy Groom, Karol Hausman, Brian Ichter, Szymon Jakubczak, Tim Jones, Liyiming Ke, Sergey Levine, Adrian Li-Bell, Mohith Mothukuri, Suraj Nair, Karl Pertsch, Lucy~Xiaoyang Shi, James Tanner, Quan Vuong, Anna Walling, Haohuan Wang, and Ury Zhilinsky.
\newblock $\pi_0$: A vision-language-action flow model for general robot control, 2024{\natexlab{b}}.
\newblock URL \url{https://arxiv.org/abs/2410.24164}.

\bibitem[Brohan et~al.(2023{\natexlab{a}})Brohan, Brown, Carbajal, Chebotar, Chen, Choromanski, Ding, Driess, Dubey, Finn, Florence, Fu, Arenas, Gopalakrishnan, Han, Hausman, Herzog, Hsu, Ichter, Irpan, Joshi, Julian, Kalashnikov, Kuang, Leal, Lee, Lee, Levine, Lu, Michalewski, Mordatch, Pertsch, Rao, Reymann, Ryoo, Salazar, Sanketi, Sermanet, Singh, Singh, Soricut, Tran, Vanhoucke, Vuong, Wahid, Welker, Wohlhart, Wu, Xia, Xiao, Xu, Xu, Yu, and Zitkovich]{brohanRT2VisionLanguageActionModels2023}
Anthony Brohan, Noah Brown, Justice Carbajal, Yevgen Chebotar, Xi~Chen, Krzysztof Choromanski, Tianli Ding, Danny Driess, Avinava Dubey, Chelsea Finn, Pete Florence, Chuyuan Fu, Montse~Gonzalez Arenas, Keerthana Gopalakrishnan, Kehang Han, Karol Hausman, Alexander Herzog, Jasmine Hsu, Brian Ichter, Alex Irpan, Nikhil Joshi, Ryan Julian, Dmitry Kalashnikov, Yuheng Kuang, Isabel Leal, Lisa Lee, Tsang-Wei~Edward Lee, Sergey Levine, Yao Lu, Henryk Michalewski, Igor Mordatch, Karl Pertsch, Kanishka Rao, Krista Reymann, Michael Ryoo, Grecia Salazar, Pannag Sanketi, Pierre Sermanet, Jaspiar Singh, Anikait Singh, Radu Soricut, Huong Tran, Vincent Vanhoucke, Quan Vuong, Ayzaan Wahid, Stefan Welker, Paul Wohlhart, Jialin Wu, Fei Xia, Ted Xiao, Peng Xu, Sichun Xu, Tianhe Yu, and Brianna Zitkovich.
\newblock {{RT-2}}: {{Vision-Language-Action Models Transfer Web Knowledge}} to {{Robotic Control}}.
\newblock \emph{arXiv preprint arXiv:2307.15818}, July 2023{\natexlab{a}}.

\bibitem[Brohan et~al.(2023{\natexlab{b}})Brohan, Brown, Carbajal, Chebotar, Dabis, Finn, Gopalakrishnan, Hausman, Herzog, Hsu, Ibarz, Ichter, Irpan, Jackson, Jesmonth, Joshi, Julian, Kalashnikov, Kuang, Leal, Lee, Levine, Lu, Malla, Manjunath, Mordatch, Nachum, Parada, Peralta, Perez, Pertsch, Quiambao, Rao, Ryoo, Salazar, Sanketi, Sayed, Singh, Sontakke, Stone, Tan, Tran, Vanhoucke, Vega, Vuong, Xia, Xiao, Xu, Xu, Yu, and Zitkovich]{brohanRT1RoboticsTransformer2023}
Anthony Brohan, Noah Brown, Justice Carbajal, Yevgen Chebotar, Joseph Dabis, Chelsea Finn, Keerthana Gopalakrishnan, Karol Hausman, Alexander Herzog, Jasmine Hsu, Julian Ibarz, Brian Ichter, Alex Irpan, Tomas Jackson, Sally Jesmonth, Nikhil Joshi, Ryan Julian, Dmitry Kalashnikov, Yuheng Kuang, Isabel Leal, Kuang-Huei Lee, Sergey Levine, Yao Lu, Utsav Malla, Deeksha Manjunath, Igor Mordatch, Ofir Nachum, Carolina Parada, Jodilyn Peralta, Emily Perez, Karl Pertsch, Jornell Quiambao, Kanishka Rao, Michael Ryoo, Grecia Salazar, Pannag Sanketi, Kevin Sayed, Jaspiar Singh, Sumedh Sontakke, Austin Stone, Clayton Tan, Huong Tran, Vincent Vanhoucke, Steve Vega, Quan Vuong, Fei Xia, Ted Xiao, Peng Xu, Sichun Xu, Tianhe Yu, and Brianna Zitkovich.
\newblock {{RT-1}}: {{Robotics Transformer}} for {{Real-World Control}} at {{Scale}}.
\newblock In \emph{Robotics: {{Science}} and {{Systems XIX}}}. {Robotics: Science and Systems Foundation}, July 2023{\natexlab{b}}.
\newblock ISBN 978-0-9923747-9-2.

\bibitem[Chi et~al.(2023)Chi, Feng, Du, Xu, Cousineau, Burchfiel, and Song]{chiDiffusionPolicyVisuomotor2023}
Cheng Chi, Siyuan Feng, Yilun Du, Zhenjia Xu, Eric Cousineau, Benjamin Burchfiel, and Shuran Song.
\newblock Diffusion {{Policy}}: {{Visuomotor Policy Learning}} via {{Action Diffusion}}.
\newblock In \emph{Robotics: {{Science}} and {{Systems XIX}}}. {Robotics: Science and Systems Foundation}, July 2023.
\newblock ISBN 978-0-9923747-9-2.

\bibitem[Cobbe et~al.(2021)Cobbe, Kosaraju, Bavarian, Chen, Jun, Kaiser, Plappert, Tworek, Hilton, Nakano, Hesse, and Schulman]{cobbeTrainingVerifiersSolve2021}
Karl Cobbe, Vineet Kosaraju, Mohammad Bavarian, Mark Chen, Heewoo Jun, Lukasz Kaiser, Matthias Plappert, Jerry Tworek, Jacob Hilton, Reiichiro Nakano, Christopher Hesse, and John Schulman.
\newblock Training {{Verifiers}} to {{Solve Math Word Problems}}.
\newblock \emph{arXiv preprint arXiv:2110.14168}, November 2021.

\bibitem[{de Haan} et~al.(2019){de Haan}, Jayaraman, and Levine]{dehaanCausalConfusionImitation2019}
Pim {de Haan}, Dinesh Jayaraman, and Sergey Levine.
\newblock Causal {{Confusion}} in {{Imitation Learning}}.
\newblock In \emph{Advances in {{Neural Information Processing Systems}}}, volume~32. Curran Associates, Inc., 2019.

\bibitem[Fu et~al.(2024)Fu, Huang, Datta, Chen, Panitch, Liu, Li, and Goldberg]{fuInContextImitationLearning2024}
Letian Fu, Huang Huang, Gaurav Datta, Lawrence~Yunliang Chen, William Chung-Ho Panitch, Fangchen Liu, Hui Li, and Ken Goldberg.
\newblock In-{{Context Imitation Learning}} via {{Next-Token Prediction}}.
\newblock \emph{arXiv preprint arXiv:2408.15980}, September 2024.

\bibitem[Haldar et~al.(2024)Haldar, Peng, and Pinto]{haldarBAKUEfficientTransformer2024}
Siddhant Haldar, Zhuoran Peng, and Lerrel Pinto.
\newblock {{BAKU}}: {{An Efficient Transformer}} for {{Multi-Task Policy Learning}}.
\newblock In \emph{The {{Thirty-eighth Annual Conference}} on {{Neural Information Processing Systems}}}, November 2024.

\bibitem[Khazatsky et~al.(2024)Khazatsky, Pertsch, Nair, Balakrishna, Dasari, Karamcheti, Nasiriany, Srirama, Chen, Ellis, Fagan, Hejna, Itkina, Lepert, Ma, Miller, Wu, Belkhale, Dass, Ha, Jain, Lee, Lee, Memmel, Park, Radosavovic, Wang, Zhan, Black, Chi, Hatch, Lin, Lu, Mercat, Rehman, Sanketi, Sharma, Simpson, Vuong, Walke, Wulfe, Xiao, Yang, Yavary, Zhao, Agia, Baijal, Castro, Chen, Chen, Chung, Drake, Foster, Gao, Herrera, Heo, Hsu, Hu, Jackson, Le, Li, Lin, Lin, Ma, Maddukuri, Mirchandani, Morton, Nguyen, O'Neill, Scalise, Seale, Son, Tian, Tran, Wang, Wu, Xie, Yang, Yin, Zhang, Bastani, Berseth, Bohg, Goldberg, Gupta, Gupta, Jayaraman, Lim, Malik, {Mart{\'i}n-Mart{\'i}n}, Ramamoorthy, Sadigh, Song, Wu, Yip, Zhu, Kollar, Levine, and Finn]{khazatskyDROIDLargeScaleInTheWild2024}
Alexander Khazatsky, Karl Pertsch, Suraj Nair, Ashwin Balakrishna, Sudeep Dasari, Siddharth Karamcheti, Soroush Nasiriany, Mohan~Kumar Srirama, Lawrence~Yunliang Chen, Kirsty Ellis, Peter~David Fagan, Joey Hejna, Masha Itkina, Marion Lepert, Yecheng~Jason Ma, Patrick~Tree Miller, Jimmy Wu, Suneel Belkhale, Shivin Dass, Huy Ha, Arhan Jain, Abraham Lee, Youngwoon Lee, Marius Memmel, Sungjae Park, Ilija Radosavovic, Kaiyuan Wang, Albert Zhan, Kevin Black, Cheng Chi, Kyle~Beltran Hatch, Shan Lin, Jingpei Lu, Jean Mercat, Abdul Rehman, Pannag~R. Sanketi, Archit Sharma, Cody Simpson, Quan Vuong, Homer~Rich Walke, Blake Wulfe, Ted Xiao, Jonathan~Heewon Yang, Arefeh Yavary, Tony~Z. Zhao, Christopher Agia, Rohan Baijal, Mateo~Guaman Castro, Daphne Chen, Qiuyu Chen, Trinity Chung, Jaimyn Drake, Ethan~Paul Foster, Jensen Gao, David~Antonio Herrera, Minho Heo, Kyle Hsu, Jiaheng Hu, Donovon Jackson, Charlotte Le, Yunshuang Li, Kevin Lin, Roy Lin, Zehan Ma, Abhiram Maddukuri, Suvir Mirchandani, Daniel Morton, Tony Nguyen,
  Abigail O'Neill, Rosario Scalise, Derick Seale, Victor Son, Stephen Tian, Emi Tran, Andrew~E. Wang, Yilin Wu, Annie Xie, Jingyun Yang, Patrick Yin, Yunchu Zhang, Osbert Bastani, Glen Berseth, Jeannette Bohg, Ken Goldberg, Abhinav Gupta, Abhishek Gupta, Dinesh Jayaraman, Joseph~J. Lim, Jitendra Malik, Roberto {Mart{\'i}n-Mart{\'i}n}, Subramanian Ramamoorthy, Dorsa Sadigh, Shuran Song, Jiajun Wu, Michael~C. Yip, Yuke Zhu, Thomas Kollar, Sergey Levine, and Chelsea Finn.
\newblock {{DROID}}: {{A Large-Scale In-The-Wild Robot Manipulation Dataset}}.
\newblock \emph{arXiv preprint arXiv:2403.12945}, March 2024.

\bibitem[Kim et~al.(2024)Kim, Pertsch, Karamcheti, Xiao, Balakrishna, Nair, Rafailov, Foster, Sanketi, Vuong, Kollar, Burchfiel, Tedrake, Sadigh, Levine, Liang, and Finn]{kimOpenVLAOpenSourceVisionLanguageAction2024}
Moo~Jin Kim, Karl Pertsch, Siddharth Karamcheti, Ted Xiao, Ashwin Balakrishna, Suraj Nair, Rafael Rafailov, Ethan~P. Foster, Pannag~R. Sanketi, Quan Vuong, Thomas Kollar, Benjamin Burchfiel, Russ Tedrake, Dorsa Sadigh, Sergey Levine, Percy Liang, and Chelsea Finn.
\newblock {{OpenVLA}}: {{An Open-Source Vision-Language-Action Model}}.
\newblock In \emph{8th {{Annual Conference}} on {{Robot Learning}}}, September 2024.

\bibitem[Lee et~al.(2024)Lee, Wang, Etukuru, Kim, Shafiullah, and Pinto]{leeBehaviorGenerationLatent2024}
Seungjae Lee, Yibin Wang, Haritheja Etukuru, H.~Jin Kim, Nur Muhammad~Mahi Shafiullah, and Lerrel Pinto.
\newblock Behavior {{Generation}} with {{Latent Actions}}.
\newblock \emph{arXiv preprint arXiv:2403.03181}, March 2024.

\bibitem[Li et~al.(2024{\natexlab{a}})Li, Li, Liu, Wang, Liu, Kang, Ma, Kong, Zhang, and Liu]{liGeneralistRobotPolicies2024}
Xinghang Li, Peiyan Li, Minghuan Liu, Dong Wang, Jirong Liu, Bingyi Kang, Xiao Ma, Tao Kong, Hanbo Zhang, and Huaping Liu.
\newblock Towards {{Generalist Robot Policies}}: {{What Matters}} in {{Building Vision-Language-Action Models}}.
\newblock \emph{arXiv preprint arXiv:2412.14058}, December 2024{\natexlab{a}}.

\bibitem[Li et~al.(2024{\natexlab{b}})Li, Deng, Zhang, Jang, Memmel, Garrett, Ramos, Fox, Li, Gupta, and Goyal]{liHAMSTERHierarchicalAction2024}
Yi~Li, Yuquan Deng, Jesse Zhang, Joel Jang, Marius Memmel, Caelan~Reed Garrett, Fabio Ramos, Dieter Fox, Anqi Li, Abhishek Gupta, and Ankit Goyal.
\newblock {{HAMSTER}}: {{Hierarchical Action Models}} for {{Open-World Robot Manipulation}}.
\newblock In \emph{The {{Thirteenth International Conference}} on {{Learning Representations}}}, October 2024{\natexlab{b}}.

\bibitem[Lightman et~al.(2023)Lightman, Kosaraju, Burda, Edwards, Baker, Lee, Leike, Schulman, Sutskever, and Cobbe]{lightmanLetsVerifyStep2023}
Hunter Lightman, Vineet Kosaraju, Yuri Burda, Harrison Edwards, Bowen Baker, Teddy Lee, Jan Leike, John Schulman, Ilya Sutskever, and Karl Cobbe.
\newblock Let's {{Verify Step}} by {{Step}}.
\newblock In \emph{The {{Twelfth International Conference}} on {{Learning Representations}}}, October 2023.

\bibitem[Liu et~al.(2024)Liu, Hamid, Xie, Lee, Du, and Finn]{liuBidirectionalDecodingImproving2024}
Yuejiang Liu, Jubayer~Ibn Hamid, Annie Xie, Yoonho Lee, Maximilian Du, and Chelsea Finn.
\newblock Bidirectional {{Decoding}}: {{Improving Action Chunking}} via {{Closed-Loop Resampling}}.
\newblock \emph{arXiv preprint arXiv:2408.17355}, December 2024.

\bibitem[Ma et~al.(2025)Ma, Tong, Jia, Hu, Su, Zhang, Yang, Li, Jaakkola, Jia, and Xie]{maInferenceTimeScalingDiffusion2025}
Nanye Ma, Shangyuan Tong, Haolin Jia, Hexiang Hu, Yu-Chuan Su, Mingda Zhang, Xuan Yang, Yandong Li, Tommi Jaakkola, Xuhui Jia, and Saining Xie.
\newblock Inference-{{Time Scaling}} for {{Diffusion Models}} beyond {{Scaling Denoising Steps}}.
\newblock \emph{arXiv preprint arXiv:2501.09732}, January 2025.

\bibitem[Mandlekar et~al.(2022)Mandlekar, Xu, Wong, Nasiriany, Wang, Kulkarni, {Fei-Fei}, Savarese, Zhu, and {Mart{\'i}n-Mart{\'i}n}]{mandlekarWhatMattersLearning2022}
Ajay Mandlekar, Danfei Xu, Josiah Wong, Soroush Nasiriany, Chen Wang, Rohun Kulkarni, Li~{Fei-Fei}, Silvio Savarese, Yuke Zhu, and Roberto {Mart{\'i}n-Mart{\'i}n}.
\newblock What {{Matters}} in {{Learning}} from {{Offline Human Demonstrations}} for {{Robot Manipulation}}.
\newblock In \emph{Proceedings of the 5th {{Conference}} on {{Robot Learning}}}, pages 1678--1690. PMLR, January 2022.

\bibitem[Nakamoto et~al.(2024)Nakamoto, Mees, Kumar, and Levine]{nakamotoSteeringYourGeneralists2024}
Mitsuhiko Nakamoto, Oier Mees, Aviral Kumar, and Sergey Levine.
\newblock Steering {{Your Generalists}}: {{Improving Robotic Foundation Models}} via {{Value Guidance}}.
\newblock In \emph{8th {{Annual Conference}} on {{Robot Learning}}}, September 2024.

\bibitem[Nasiriany et~al.(2024{\natexlab{a}})Nasiriany, Kirmani, Ding, Smith, Zhu, Driess, Sadigh, and Xiao]{rt-affordance}
Soroush Nasiriany, Sean Kirmani, Tianli Ding, Laura Smith, Yuke Zhu, Danny Driess, Dorsa Sadigh, and Ted Xiao.
\newblock Rt-affordance: Affordances are versatile intermediate representations for robot manipulation, 2024{\natexlab{a}}.
\newblock URL \url{https://arxiv.org/abs/2411.02704}.

\bibitem[Nasiriany et~al.(2024{\natexlab{b}})Nasiriany, Maddukuri, Zhang, Parikh, Lo, Joshi, Mandlekar, and Zhu]{nasirianyRoboCasaLargeScaleSimulation2024}
Soroush Nasiriany, Abhiram Maddukuri, Lance Zhang, Adeet Parikh, Aaron Lo, Abhishek Joshi, Ajay Mandlekar, and Yuke Zhu.
\newblock {{RoboCasa}}: {{Large-Scale Simulation}} of {{Everyday Tasks}} for {{Generalist Robots}}.
\newblock \emph{arXiv preprint arXiv:2406.02523}, June 2024{\natexlab{b}}.

\bibitem[Radosavovic et~al.(2023)Radosavovic, Shi, Fu, Goldberg, Darrell, and Malik]{radosavovicRobotLearningSensorimotor2023}
Ilija Radosavovic, Baifeng Shi, Letian Fu, Ken Goldberg, Trevor Darrell, and Jitendra Malik.
\newblock Robot {{Learning}} with {{Sensorimotor Pre-training}}.
\newblock In \emph{Proceedings of {{The}} 7th {{Conference}} on {{Robot Learning}}}, pages 683--693. PMLR, December 2023.

\bibitem[Ravichandar et~al.(2020)Ravichandar, Polydoros, Chernova, and Billard]{ravichandarRecentAdvancesRobot2020}
Harish Ravichandar, Athanasios~S. Polydoros, Sonia Chernova, and Aude Billard.
\newblock Recent {{Advances}} in {{Robot Learning}} from {{Demonstration}}.
\newblock \emph{Annual Review of Control, Robotics, and Autonomous Systems}, 3\penalty0 (Volume 3, 2020):\penalty0 297--330, May 2020.

\bibitem[Ren et~al.(2025)Ren, Sundaresan, Sadigh, Choudhury, and Bohg]{motiontracksil}
Juntao Ren, Priya Sundaresan, Dorsa Sadigh, Sanjiban Choudhury, and Jeannette Bohg.
\newblock Motion tracks: A unified representation for human-robot transfer in few-shot imitation learning, 2025.
\newblock URL \url{https://arxiv.org/abs/2501.06994}.

\bibitem[Ross and Bagnell(2010)]{rossEfficientReductionsImitation2010a}
Stephane Ross and Drew Bagnell.
\newblock Efficient {{Reductions}} for {{Imitation Learning}}.
\newblock In \emph{Proceedings of the {{Thirteenth International Conference}} on {{Artificial Intelligence}} and {{Statistics}}}, pages 661--668. {JMLR Workshop and Conference Proceedings}, March 2010.

\bibitem[Ross et~al.(2011)Ross, Gordon, and Bagnell]{dagger}
Stephane Ross, Geoffrey~J. Gordon, and J.~Andrew Bagnell.
\newblock A reduction of imitation learning and structured prediction to no-regret online learning, 2011.
\newblock URL \url{https://arxiv.org/abs/1011.0686}.

\bibitem[Seo et~al.(2023{\natexlab{a}})Seo, Hwang, Yang, and Kim]{PALR}
Seokin Seo, HyeongJoo Hwang, Hongseok Yang, and Kee-Eung Kim.
\newblock Regularized behavior cloning for blocking the leakage of past action information.
\newblock In A.~Oh, T.~Naumann, A.~Globerson, K.~Saenko, M.~Hardt, and S.~Levine, editors, \emph{Advances in Neural Information Processing Systems}, volume~36, pages 2128--2153. Curran Associates, Inc., 2023{\natexlab{a}}.
\newblock URL \url{https://proceedings.neurips.cc/paper_files/paper/2023/file/06b71ad997f7e3e4b2e2f2ea12e5a759-Paper-Conference.pdf}.

\bibitem[Seo et~al.(2023{\natexlab{b}})Seo, Hwang, Yang, and Kim]{seoRegularizedBehaviorCloning2023}
Seokin Seo, HyeongJoo Hwang, Hongseok Yang, and Kee-Eung Kim.
\newblock Regularized {{Behavior Cloning}} for {{Blocking}} the {{Leakage}} of {{Past Action Information}}.
\newblock \emph{Advances in Neural Information Processing Systems}, 36:\penalty0 2128--2153, December 2023{\natexlab{b}}.

\bibitem[Shao et~al.(2025)Shao, Buening, and Kwiatkowska]{shaoUnifyingFrameworkCausal2025}
Daqian Shao, Thomas~Kleine Buening, and Marta Kwiatkowska.
\newblock A {{Unifying Framework}} for {{Causal Imitation Learning}} with {{Hidden Confounders}}.
\newblock \emph{arXiv preprint arXiv:2502.07656}, February 2025.

\bibitem[Snell et~al.(2024)Snell, Lee, Xu, and Kumar]{llmscaling}
Charlie Snell, Jaehoon Lee, Kelvin Xu, and Aviral Kumar.
\newblock Scaling llm test-time compute optimally can be more effective than scaling model parameters, 2024.
\newblock URL \url{https://arxiv.org/abs/2408.03314}.

\bibitem[Spencer et~al.(2021)Spencer, Choudhury, Venkatraman, Ziebart, and Bagnell]{ILfeedback}
Jonathan Spencer, Sanjiban Choudhury, Arun Venkatraman, Brian Ziebart, and J.~Andrew Bagnell.
\newblock Feedback in imitation learning: The three regimes of covariate shift, 2021.
\newblock URL \url{https://arxiv.org/abs/2102.02872}.

\bibitem[Stechly et~al.(2024)Stechly, Valmeekam, and Kambhampati]{stechlySelfVerificationLimitationsLarge2024}
Kaya Stechly, Karthik Valmeekam, and Subbarao Kambhampati.
\newblock On the {{Self-Verification Limitations}} of {{Large Language Models}} on {{Reasoning}} and {{Planning Tasks}}.
\newblock \emph{arXiv preprint arXiv:2402.08115}, August 2024.

\bibitem[Sundaresan et~al.(2024)Sundaresan, Vuong, Gu, Xu, Xiao, Kirmani, Yu, Stark, Jain, Hausman, Sadigh, Bohg, and Schaal]{rt-sketch}
Priya Sundaresan, Quan Vuong, Jiayuan Gu, Peng Xu, Ted Xiao, Sean Kirmani, Tianhe Yu, Michael Stark, Ajinkya Jain, Karol Hausman, Dorsa Sadigh, Jeannette Bohg, and Stefan Schaal.
\newblock Rt-sketch: Goal-conditioned imitation learning from hand-drawn sketches, 2024.
\newblock URL \url{https://arxiv.org/abs/2403.02709}.

\bibitem[Team et~al.(2024)Team, Ghosh, Walke, Pertsch, Black, Mees, Dasari, Hejna, Kreiman, Xu, Luo, Tan, Chen, Sanketi, Vuong, Xiao, Sadigh, Finn, and Levine]{teamOctoOpenSourceGeneralist2024}
Octo~Model Team, Dibya Ghosh, Homer Walke, Karl Pertsch, Kevin Black, Oier Mees, Sudeep Dasari, Joey Hejna, Tobias Kreiman, Charles Xu, Jianlan Luo, You~Liang Tan, Lawrence~Yunliang Chen, Pannag Sanketi, Quan Vuong, Ted Xiao, Dorsa Sadigh, Chelsea Finn, and Sergey Levine.
\newblock Octo: {{An Open-Source Generalist Robot Policy}}.
\newblock \emph{arXiv preprint arXiv:2405.12213}, May 2024.

\bibitem[Torne et~al.(2024)Torne, Jain, Yuan, Macha, Ankile, Simeonov, Agrawal, and Gupta]{torneRobotLearningSuperLinear2024}
Marcel Torne, Arhan Jain, Jiayi Yuan, Vidaaranya Macha, Lars Ankile, Anthony Simeonov, Pulkit Agrawal, and Abhishek Gupta.
\newblock Robot {{Learning}} with {{Super-Linear Scaling}}.
\newblock \emph{arXiv preprint arXiv:2412.01770}, December 2024.

\bibitem[Vuong et~al.(2025)Vuong, Vu, An, and Reid]{vuongActionTokenizerMatters2025}
An~Dinh Vuong, Minh~Nhat Vu, Dong An, and Ian Reid.
\newblock Action {{Tokenizer Matters}} in {{In-Context Imitation Learning}}.
\newblock \emph{arXiv preprint arXiv:2503.01206}, March 2025.

\bibitem[Wang et~al.(2024)Wang, Hart, Surovik, Kelestemur, Huang, Zhao, Yeatman, Wang, Walters, and Platt]{wangEquivariantDiffusionPolicy2024}
Dian Wang, Stephen Hart, David Surovik, Tarik Kelestemur, Haojie Huang, Haibo Zhao, Mark Yeatman, Jiuguang Wang, Robin Walters, and Robert Platt.
\newblock Equivariant {{Diffusion Policy}}.
\newblock In \emph{8th {{Annual Conference}} on {{Robot Learning}}}, September 2024.

\bibitem[Wen et~al.(2020)Wen, Lin, Darrell, Jayaraman, and Gao]{wenFightingCopycatAgents2020}
Chuan Wen, Jierui Lin, Trevor Darrell, Dinesh Jayaraman, and Yang Gao.
\newblock Fighting {{Copycat Agents}} in {{Behavioral Cloning}} from {{Observation Histories}}.
\newblock In \emph{Advances in {{Neural Information Processing Systems}}}, volume~33, pages 2564--2575. Curran Associates, Inc., 2020.

\bibitem[Wen et~al.(2021)Wen, Lin, Qian, Gao, and Jayaraman]{keyframeil}
Chuan Wen, Jierui Lin, Jianing Qian, Yang Gao, and Dinesh Jayaraman.
\newblock Keyframe-focused visual imitation learning, 2021.
\newblock URL \url{https://arxiv.org/abs/2106.06452}.

\bibitem[Weng et~al.(2023)Weng, Zhu, Xia, Li, He, Liu, Sun, Liu, and Zhao]{wengLargeLanguageModels2023}
Yixuan Weng, Minjun Zhu, Fei Xia, Bin Li, Shizhu He, Shengping Liu, Bin Sun, Kang Liu, and Jun Zhao.
\newblock Large {{Language Models}} are {{Better Reasoners}} with {{Self-Verification}}.
\newblock In \emph{The 2023 {{Conference}} on {{Empirical Methods}} in {{Natural Language Processing}}}, December 2023.

\bibitem[Yu et~al.(2024)Yu, Gao, and Wang]{yuOVMOutcomesupervisedValue2024}
Fei Yu, Anningzhe Gao, and Benyou Wang.
\newblock {{OVM}}, {{Outcome-supervised Value Models}} for {{Planning}} in {{Mathematical Reasoning}}.
\newblock In Kevin Duh, Helena Gomez, and Steven Bethard, editors, \emph{Findings of the {{Association}} for {{Computational Linguistics}}: {{NAACL}} 2024}, pages 858--875, Mexico City, Mexico, June 2024. Association for Computational Linguistics.

\bibitem[Zare et~al.(2024)Zare, Kebria, Khosravi, and Nahavandi]{zareSurveyImitationLearning2024}
Maryam Zare, Parham~M. Kebria, Abbas Khosravi, and Saeid Nahavandi.
\newblock A {{Survey}} of {{Imitation Learning}}: {{Algorithms}}, {{Recent Developments}}, and {{Challenges}}.
\newblock \emph{IEEE Transactions on Cybernetics}, 54\penalty0 (12):\penalty0 7173--7186, December 2024.

\bibitem[Ze et~al.(2024)Ze, Zhang, Zhang, Hu, Wang, and Xu]{ze3DDiffusionPolicy2024}
Yanjie Ze, Gu~Zhang, Kangning Zhang, Chenyuan Hu, Muhan Wang, and Huazhe Xu.
\newblock {{3D Diffusion Policy}}.
\newblock \emph{arXiv preprint arXiv:2403.03954}, March 2024.

\bibitem[Zhao et~al.(2023)Zhao, Kumar, Levine, and Finn]{zhaoLearningFineGrainedBimanual2023}
Tony~Z. Zhao, Vikash Kumar, Sergey Levine, and Chelsea Finn.
\newblock Learning {{Fine-Grained Bimanual Manipulation}} with {{Low-Cost Hardware}}.
\newblock \emph{arXiv preprint arXiv:2304.13705}, April 2023.

\bibitem[Zheng et~al.(2024)Zheng, Liang, Huang, Gao, III, Kolobov, Huang, and Yang]{traceVLA}
Ruijie Zheng, Yongyuan Liang, Shuaiyi Huang, Jianfeng Gao, Hal~Daumé III, Andrey Kolobov, Furong Huang, and Jianwei Yang.
\newblock Tracevla: Visual trace prompting enhances spatial-temporal awareness for generalist robotic policies, 2024.
\newblock URL \url{https://arxiv.org/abs/2412.10345}.

\end{thebibliography}

\newpage
\appendix
\label{sec:appdx}

\section{Additional Experiments}
\label{apdx:experiment}

In addition to the main results presented in~\cref{sec:simexp}, we conduct three experiments to further validate the design decisions behind our proposed method.

\subsection{Which component of the policy benefits most from PTP?}
\label{apdx:effectinternal}

\begin{figure}[h!]
    \centering
    \includegraphics[width=0.3\linewidth]{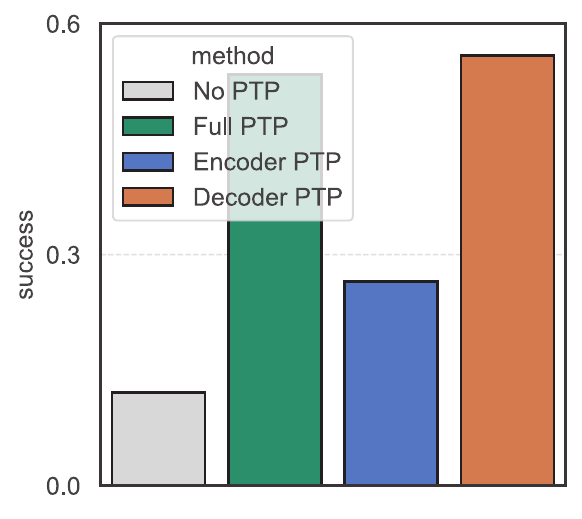}
    \caption{Performance of ablated PTP variants on Push-T. Applying PTP only to the decoder recovers the full PTP policy performance, whereas encoder-only PTP does not.}
    \label{fig:appendixanalysis}
\end{figure}

To identify which part of the policy is most influenced by PTP, we compare a fully PTP-trained long-context policy against two ablated variants: {\it Encoder PTP}, where we first train the visual encoder with PTP, then freeze it and train the action decoder without PTP; {\it Decoder PTP}, where we conversely train the encoder without PTP, freeze it, and then apply PTP only during decoder training.
As shown in~\cref{fig:appendixanalysis}, {\it Decoder PTP} achieves performance on par with the fully trained PTP policy, whereas {\it Encoder PTP} performs significantly worse. This result suggests that the benefits of PTP primarily stem from improved temporal modeling in the action decoder, rather than from changes to the visual encoder, directly motivating our multi-stage training recipe that decouples encoder pretraining from long-context policy learning.

\subsection{Does our method reduce reliance on action chunking?}
\label{apdx:chunking}

\begin{figure}[h!]
    \centering
    \includegraphics[width=\linewidth]{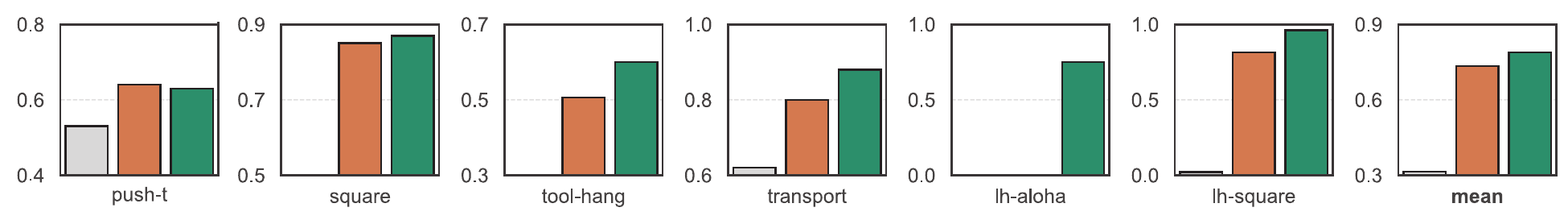}
    \newline
    \includegraphics[width=0.4\linewidth]{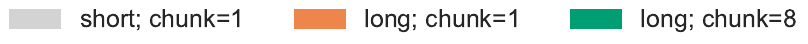}
    \caption{
    Comparison of policies with different context lengths and chunk sizes.
    Long-context policies trained with PTP perform significantly better than short-context policies when run in a fully closed-loop setting (chunk size 1). Moreover, they achieve performance comparable to long-context policies that use open-loop chunking (chunk size 8), indicating reduced reliance on chunking during execution.
    }
    \label{fig:closedloopres}
\end{figure}

Existing short-context policies typically rely on action chunking compensate for limited access to past observations. However, this common design choice comes at the cost of reduced reactivity.
To assess whether our method alleviates this limitation, we compare the performance of three policy variants:
(i) {\it short-context short-chunk}, which receives the past 2 frames as input and outputs single-step actions (chunk size 1);
(ii) {\it long-context short-chunk}, which receives the past 16 frames and also outputs single-step actions;
and (iii) {\it long-context long-chunk}, which receives the past 16 frames and outputs action chunks of size 8.
As shown in~\cref{fig:closedloopres}, the {\it long-context short-chunk} policies trained by our method substantially outperform the {\it short-context} counterparts and recover most of the performance of the {\it long-context long-chunk} policies.
This result demonstrates the effectiveness of our method in reducing reliance on open-loop action chunking.

\subsection{Is PTP still critical when conditioning on past actions?}

\begin{figure}[h!]
    \centering
    \includegraphics[width=0.9\linewidth]{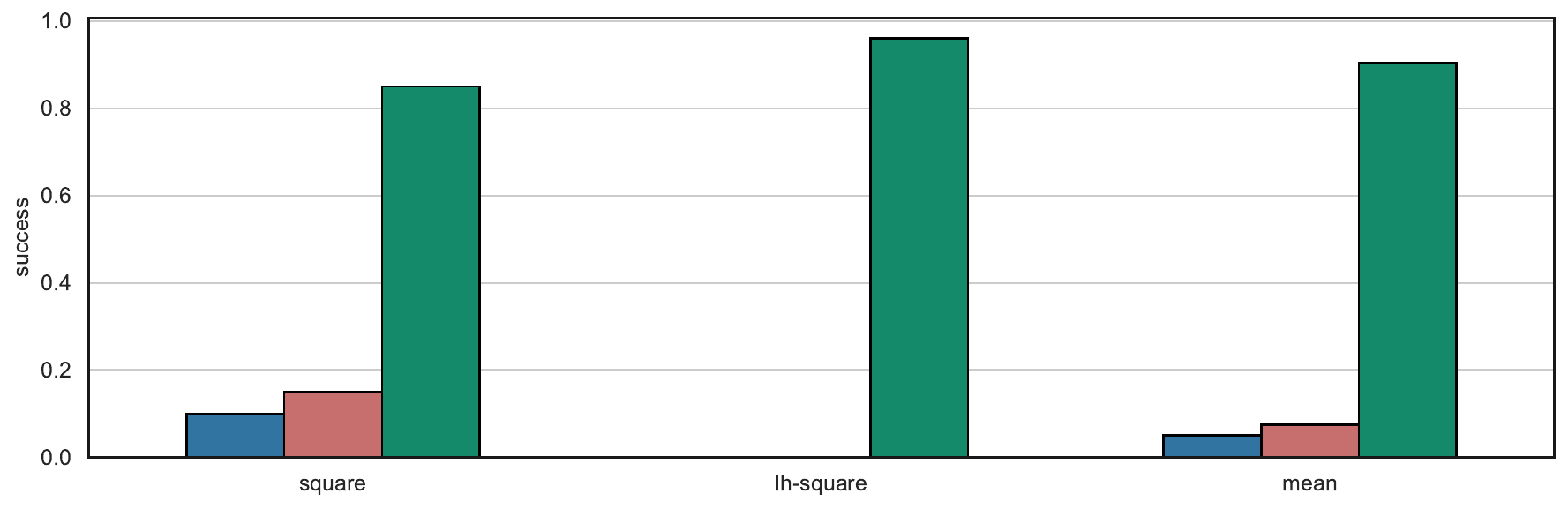}
    \includegraphics[width=0.4\linewidth]{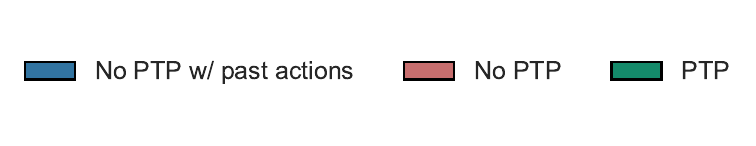}
    \caption{Comparison of adding past actions into the context history without PTP and our baseline of PTP and no PTP without actions. We observe that adding past actions in the observation doesn't improve performance.}
    \label{fig:pastaction_figure}
\end{figure}

Our earlier analysis in~\cref{fig:ptp_correlation} has shown the importance of PTP in capturing temporal action dependencies when the policy is conditioned on past observations. A natural question is whether PTP remains necessary when the model also has direct access to past actions.
To understand this, we augment the input to diffusion policies with the previous 16 actions and compare performance with and without PTP. As shown in~\cref{fig:pastaction_figure}, even with access to past actions, the vanilla baseline performs poorly without PTP, while our method consistently yields substantially better results. Consistent with our previous findings, this result highlights the critical role of PTP in enabling diffusion policies to effectively model temporal structure, even when past actions are explicitly provided.

\section{Environmental Details}
\label{apdx:envanddata}

\subsection{Real-World Tasks}

We conduct real-world experiments using two robot platforms: a Franka Panda arm set up, and the ALOHA bimanual system.
For the Franka setup, we follow the DROID hardware configuration~\citep{khazatskyDROIDLargeScaleInTheWild2024}, using a single arm with wrist-mounted RGB cameras and proprioceptive sensing. The observation space includes RGB images and end-effector pose, while the action space consists of 6-DoF Cartesian displacements and gripper commands. We collect 50-200 human demonstrations for each task.
For the ALOHA platform, we use the bimanual robot as described in~\citep{zhaoLearningFineGrainedBimanual2023}, with RGB camera inputs and proprioceptive feedback. The observation space consists of dual-arm RGB views and proprioception, and the action space includes joint displacements for both arms and gripper states. We collect 150 demonstrations for the tape replacement task.

\subsection{Simulation Tasks}

Our simulation tasks include the Push-T task~\citep{chiDiffusionPolicyVisuomotor2023}, three existing tasks in the Robomimic benchmark~\citep{mandlekarWhatMattersLearning2022}, along with two new long-horizon tasks that we introduce:
\begin{itemize}[nosep]
    \item {\it Long-Horizon Square}: A variant of the RoboMimic square task, where the robot must place a block onto the farthest peg from its initial location. We collect 100 noisy scripted demonstrations to prevent the policy from inferring the goal using current pose information alone, thus requiring memory of the initial state.
    \item {\it Long-Horizon ALOHA}: A simulated bimanual task where one arm picks up a block, moves it to the center of the workspace, and places it back at its original location. Success requires remembering the block’s starting position, highlighting the need for long-term memory.
\end{itemize}

\begin{table}[t]
\centering
\small
\caption{Hyper-parameters in simulation and real-world experiments.}
\vskip 0.05in
\begin{tabular}{l|cccccc}
\toprule
Hyperparameter & LH Square & LH Aloha & Block Move & Two Scoops & Mug Repl. & Tape Repl. \\
\midrule
Epochs & 500 & 1500 & 500 & 500 & 500 & 1500 \\
\# Demos & 100 & 50 & 50 & 200 & 200 & 150 \\
\# Subsampled frames & 20 & 20 & 1 & 20 & 1 & 24 \\
\# Observations & 20 & 32 & 10 & 20 & 32 & 20 \\
\bottomrule
\end{tabular}
\label{tab:hyperparams}
\end{table}

\subsection{Implementation Details}

\subsubsection{Policy Architecture}
We build upon the transformer-based Diffusion Policy codebase~\cite{chiDiffusionPolicyVisuomotor2023}, which supports training and evaluation across multiple Robomimic tasks. All policies are trained for 500 epochs by default, using visual encoders and chunked action prediction.
For long-horizon ALOHA tasks, we train for 1500 epochs to accommodate the added difficulty of bimanual coordination and higher-frequency control. 
To reduce training overhead, all long-context policies are initialized with a frozen short-context encoder, pretrained on 2-frame inputs. 
This design choice is supported by the analysis in~\cref{fig:appendixanalysis}, which shows that freezing the encoder does not impair performance.
To further improve training efficiency, we cache visual embeddings during data preprocessing and load them at runtime. This avoids repeatedly passing observations through the encoder and speeds up training.

\subsubsection{Subsampling Rate}
Real-world tasks often require longer history horizons, but full-length observation sequences can be computationally expensive. To reduce inference latency, we apply temporal subsampling to the input sequence. Specifically, instead of feeding all $T$ observations ${t_0, t_1, ..., t_{T-1}}$, we sample every $K$th frame, i.e., ${t_{K-1}, t_{2K-1}, ..., t_{T-1}}$, where $T$ is a multiple of $K$.
This reduces the effective observation size while retaining broad temporal coverage. Subsampling values are listed in~\cref{tab:hyperparams}.

\subsubsection{Context Length}

When increasing the observation history, we scale the prediction horizon (past and future tokens) proportionally. We empirically find that the prediction length of future tokens has only a minor effect on task performance.
Detailed context length configurations are provided in~\cref{tab:contextlength}.

\begin{table}[h]
\caption{Settings for different context lengths}
\label{tab:contextlength}
\vskip 0.05in
\begin{center}
\begin{small}
\begin{tabular}{l|ccccccr}
\toprule
Observation Length & 2 & 4 & 8 & 16 \\
\midrule
Horizon & 16 & 20 & 24 & 32 \\
Future Tokens & 14 & 16 & 16 & 16 \\
\bottomrule
\end{tabular}
\end{small}
\end{center}
\vskip -0.1in
\end{table}

\subsubsection{Action Dependency Metric}

To quantify how well a policy captures temporal action structure, we use \emph{action predictability} as a proxy metric~\citep{wenFightingCopycatAgents2020}. Specifically, we measure how accurately the current action $a_t$ can be predicted from a window of $K=15$ past actions, defined as $p(a_t \mid a_{t-K:t-1})$.
We compute this quantity over policy rollouts and compare it to the same metric evaluated on the expert demonstrations. Higher predictability indicates stronger temporal action dependencies captured by the learned policy.

\section{Additional Results}

In addition to the results reported in~\cref{sec:simexp}, we report per-task success rates across varying temporal context lengths, training conditions, and chunking configurations.
\cref{tab:success-rates} compares diffusion-based and regression-based baselines on six benchmark tasks, evaluated under different history lengths and with or without PTP.
\cref{robomimic-nochunk} presents results in the closed-loop setting (chunk size = 1) across different context lengths.

\begin{table}[H]
\caption{Success rate (\%) of diffusion-based and regression-based policies on simulation tasks under different training and history conditions. Results are reported as mean ± standard deviation across 3 seeds.}
\label{tab:success-rates}
\vskip 0.05in
\begin{center}
\begin{small}
\begin{tabular}{l|cccccc}
\toprule
Method & Push-T & Square & Tool-Hang & Transport & ALOHA & Long Square \\
\midrule
Diffusion (PTP) & 0.62 ± 0.02 & \textbf{0.89 ± 0.01} & \textbf{0.75 ± 0.10} & \textbf{0.67 ± 0.08} & 0.98 ± 0.01& \textbf{0.93 ± 0.02} \\
Diffusion (no-PTP) & 0.59 ± 0.01 & 0.17 ± 0.01 & 0.00 ± 0.00 & 0.00 ± 0.00 & 0.20 ± 0.19 & 0.03 ± 0.02 \\
Diffusion (no-hist) & \textbf{0.67 ± 0.03} & 0.79 ± 0.06 & 0.51 ± 0.14 & 0.60 ± 0.08 & 0.28 ± 0.04 & 0.12 ± 0.05 \\
Regression (PTP) & 0.40 ± 0.10 & 0.74 ± 0.02 & 0.41 ± 0.01 & 0.63 ± 0.04 & 0.99 ± 0.02 & 0.90 ± 0.05 \\
Regression (no-PTP) & 0.31 ± 0.31 & 0.78 ± 0.03 & 0.40 ± 0.05 & 0.51 ± 0.01 & \textbf{1.00 ± 0.00} & 0.89 ± 0.01 \\
Regression (no-hist) & 0.64 ± 0.08 & 0.19 ± 0.01 & 0.14 ± 0.04 & 0.48 ± 0.03 & 0.43 ± 0.03 & 0.00 ± 0.00 \\
\bottomrule
\end{tabular}
\end{small}
\end{center}
\vskip -0.1in
\end{table}

\begin{table}[H]
\caption{Success rate (\%) on simulation tasks under closed-loop execution (chunk size = 1) .}
\label{robomimic-nochunk}
\vskip 0.05in
\begin{center}
\begin{small}
\begin{tabular}{l|ccccc|c}
\toprule
Observations & Push-T & Square & Tool Hang & Transport & Long Square & Mean \\
\midrule
2  & 0.53 & 0.13 & 0.62 & 0.053 & 0.02 & 0.37 \\
4  & 0.53 & 0.68 & 0.84 & 0.13  & 0.02 & 0.51 \\
8  & 0.59 & 0.83 & \textbf{0.86} & 0.48 & 0.11 & 0.63 \\
16 & \textbf{0.64} & \textbf{0.85} & 0.82 & \textbf{0.51} & \textbf{0.81} & \textbf{0.77} \\
\bottomrule
\end{tabular}
\end{small}
\end{center}
\vskip -0.1in
\end{table}

\end{document}